\documentclass[a4paper,fleqn]{cas-dc}
\usepackage{url}
\usepackage{multirow}
\usepackage{graphicx}
\usepackage{booktabs}      
\usepackage{multirow}      
\usepackage[table]{xcolor} 
\usepackage{threeparttable}
\usepackage{lscape}
\usepackage{caption}
\usepackage{subfig}
\usepackage{float}
\usepackage{adjustbox}
\usepackage{booktabs}
\usepackage{enumitem}
\usepackage{amssymb,amsfonts}
\usepackage{algorithm}
\usepackage{algpseudocode}
\usepackage{amsmath}
\usepackage{bbm}          
\usepackage{gensymb}

\usepackage{xcolor}
\usepackage{colortbl}
\usepackage{siunitx}

\usepackage{hyperref}
\hypersetup{
    colorlinks=true,
    linkcolor=blue,   
    citecolor=black,  
    urlcolor=blue
}

\newcommand{\mypar}[1]{\noindent\textbf{#1}~}

\def\tsc#1{\csdef{#1}{\textsc{\lowercase{#1}}\xspace}}
\tsc{WGM}
\tsc{QE}
\tsc{EP}
\tsc{PMS}
\tsc{BEC}
\tsc{DE}

\usepackage{natbib}
\usepackage{microtype}
\begin{document}

\let\WriteBookmarks\relax
\def\floatpagepagefraction{1}
\def\textpagefraction{.001}

\title [mode=title]{GMGaze: MoE-Based Context-Aware Gaze Estimation with CLIP and Multiscale Transformer}

\author{Xinyuan~Zhao$^{a}$, Yihang~Wu$^{a}$, Ahmad~Chaddad$^{a,b,*}$, Sarah A. Alkhodair$^{c}$, Reem~Kateb$^{d,e}$}

\address[add1]{School of Artificial Intelligence, Guilin University of Electronic Technology, Guilin, China}
\address[add2]{The Laboratory for Imagery, Vision and Artificial Intelligence, École de Technologie Supérieure, Montreal, Canada}
\address[add3]{Information Technology Department, College of Computer and Information Sciences, King Saud University, Riyadh, Saudi Arabia}
\address[add4]{Department of Cybersecurity, College of Computer Science and Engineering, Taibah University, Medina, Saudi Arabia}
\address[add5]{Department of Networked Engineering, College of Computer Science and Engineering, Jeddah University, Jeddah, Saudi Arabia}
\cortext[cor1]{ahmad8chaddad@gmail.com}

\fntext[fn1]{Xinyuan Zhao, Yihang Wu and Ahmad Chaddad contributed equally to this work.}


\begin{abstract}
Gaze estimation methods commonly use facial appearances to predict the direction of a person gaze. However, previous studies show three major challenges with convolutional neural network (CNN)-based, transformer-based, and contrastive language-image pre-training (CLIP)-based methods, including late fusion of image features, lack of factor-aware conditioning, and impractical capacity scaling. To address these challenges, we propose Globally-conditioned Multi-scale Gaze estimation (GMGaze), which leverages a multi-scale transformer architecture. Specifically, the model first introduces semantic prototype conditioning, which modulates the CLIP global image embedding using four learned prototype banks (i.e., illumination, background, head pose and appearance) to generate two complementary context-biased global tokens. These tokens, along with the CLIP patch and CNN tokens, are fused at the first layer. This early unified fusion prevents information loss common in late-stage merging. Finally, each token passes through sparse Mixture-of-Experts modules, providing conditional computational capacity without uniformly increasing dense parameters. For cross-domain adaptation, we incorporate an adversarial domain adaptation technique with a feature separation loss that encourages the two global tokens to remain de-correlated. Experiments using four public benchmarks (MPIIFaceGaze, EYEDIAP, Gaze360, and ETH-XGaze) show that GMGaze achieves mean angular errors of 2.49$^\circ$, 3.22$^\circ$, 10.16$^\circ$, and 1.44$^\circ$, respectively, outperforming previous baselines in all within-domain settings. In cross-domain evaluations, it provides state-of-the-art (SOTA) results on two standard transfer routes. The code is publicly available at \url{https://github.com/AIPMLab/GazeFormer-MoE}.

\end{abstract}



\maketitle

\section{Introduction}\label{int}

Appearance-based gaze estimation, which maps a monocular RGB face image to a 3D gaze direction, plays a crucial role in applications such as augmented reality, accessibility, and behavioral analysis \citep{wu2024eg,wang2016appearance,yuan2022self}. Despite many studies in controlled environments, deploying gaze estimation systems in real-world settings remains challenging due to the presence of nuisance factors, including illumination variations, head pose changes, and background clutter. These factors obscure subtle peri-ocular cues that are important for accurate gaze prediction, leading to degraded generalization performance across domains. To address these complexities, bio-inspired models like the Lobula Giant Movement Detector have been proposed to mimic human-like central fixation bias for robust sensing in open scenes \cite{xu2023bio}. Through a comprehensive review of existing literature, we identify three key challenges in current gaze estimation models. First, \textit{late fusion} limits the interaction between high-level semantic features and fine-grained visual details, as multi-scale representations are typically merged at deeper layers (e.g. deep learning models). Second, the \textit{lack of factor-aware conditioning} results in global image representations that are highly entangled, making it difficult to separate gaze-relevant information from environmental variations. Third, \textit{uniform capacity scaling} leads to inefficient computation, as model capacity is increased indiscriminately rather than being dynamically allocated to more informative or complex visual tokens.

However, these challenges are fundamentally caused by the limitations of existing architectural paradigms. For example, convolutional Neural Networks (CNNs) \citep{zhang2017s,cheng2020coarse} inherently introduce early information fusion due to their hierarchical structure, resulting in gaze-relevant cues to be entangled with identity and illumination factors. Similarly, transformer-based approaches \citep{cheng2022gaze} provide global information but lack prior-guided attention, as it allocates the same weight to all image regions, including irrelevant background areas. Recently, Contrastive Language-Image Pre-training (CLIP) based methods have introduced powerful semantic representations \citep{wang2023gazeclip}. However, their approach relies on a single global embedding, which fails to distinguish between different contextual factors. Although domain adaptation techniques \citep{xia2025collaborative,zeng2025gaze} aim to improve generalization, they operate on entangled representations, limiting their potential. Furthermore, studies in wearable eye tracking \citep{homavazir2025slippage} suggest that gaze estimators remain sensitive to real-world perturbations such as device shifts and calibration drift, highlighting the need for robust feature representations.

\begin{figure}
     \centering
\includegraphics[width=1\linewidth]{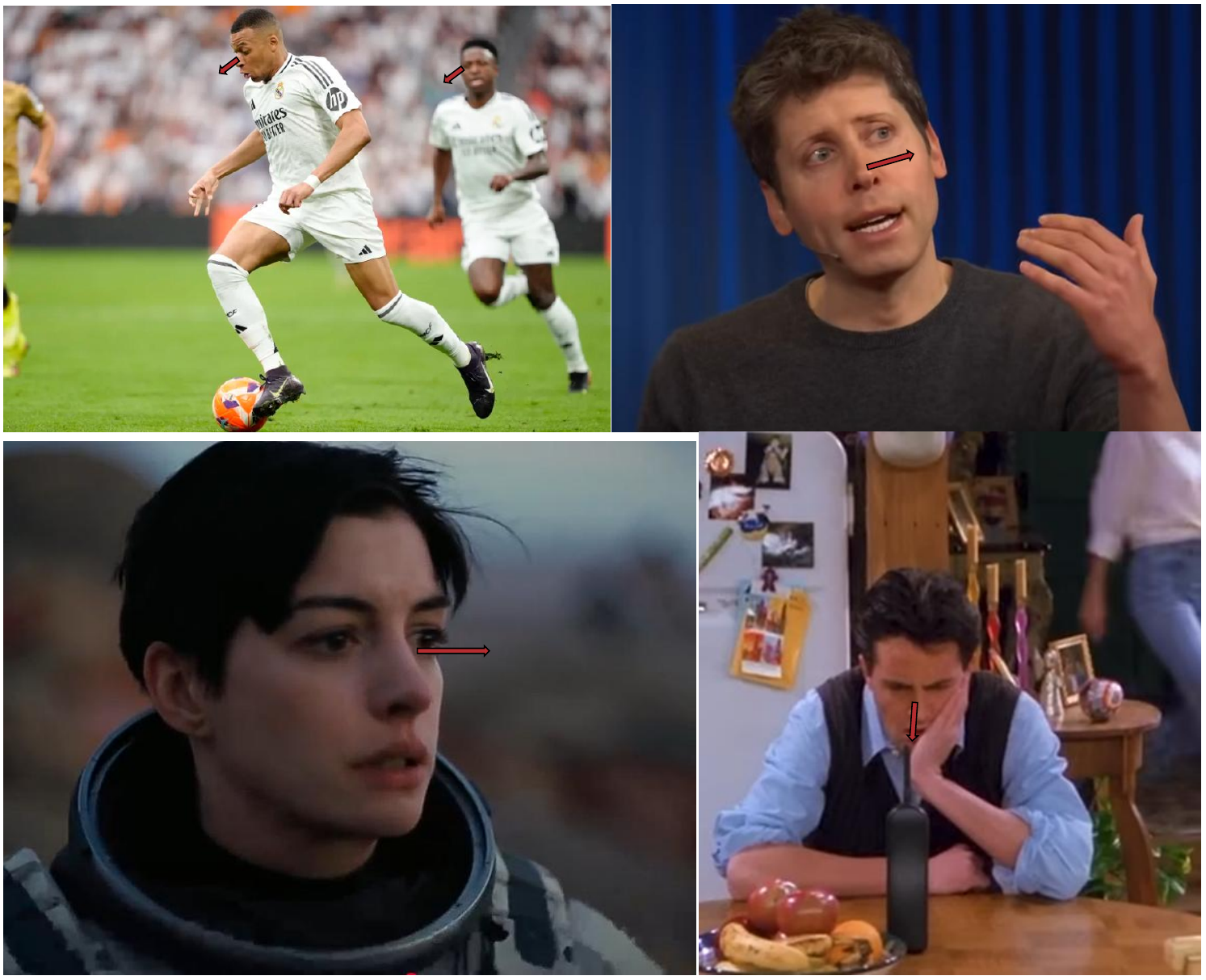}
\caption{Gaze visualization of the proposed GMGaze on unseen frames from public videos.}
\label{vis_unseened}
\end{figure}

To address these limitations, this study proposes GMGaze, a Mixture-of-Experts (MoE) based multi-scale transformer framework. Specifically, we introduce a semantic prototype conditioning mechanism that decomposes global CLIP embeddings into complementary, context-aware representations, enabling explicit factor-aware modeling. Furthermore, we design an early unified token fusion strategy that integrates global, patch-level, and CNN features at the input stage, allowing cross-scale interactions from the beginning to mitigate the late fusion.  In addition, we replace standard feed-forward networks with sparse MoE layers to achieve token-level dynamic capacity allocation, improving computational efficiency. To enhance cross-domain robustness, we incorporate an adversarial domain adaptation strategy with a feature separation loss, encouraging the learned representations to remain disentangled during training. Figure \ref{vis_unseened} illustrates gaze visualization of the proposed GMGaze on unseen frames from public videos, highlighting the potential for real-world applications. The main contributions of this work are summarized as follows:
\begin{enumerate}[leftmargin=1.5em, itemsep=1pt, topsep=2pt]
  \item We propose GMGaze, a unified multi-scale transformer framework that integrates visual and semantic representations for robust gaze estimation.
  \item We introduce a semantic prototype conditioning mechanism to decompose global representations into factor-specific components, improving interpretability and robustness.
  \item We design a token-level Mixture-of-Experts architecture for dynamic capacity scaling, enabling efficient and adaptive computation.
  \item We develop a feature separation loss combined with adversarial adaptation to enhance cross-domain generalization, achieving state-of-the-art performance on multiple public benchmarks.
\end{enumerate}

This study further extends the Gazeformer-MoE presented in our previous work \cite{zhao2026gazeformer} by introducing a feature separation loss to improve cross-domain performance. Furthermore, we include additional within-domain benchmarks, cross-domain experiments, qualitative visualizations, and comprehensive ablation studies to confirm the effectiveness of the proposed method.

The remainder of this paper is organized as follows. Section \ref{RW} summarizes the related work for gaze estimation. Section~\ref{methodology} describes the gaze estimation task and details the proposed approach. Section~\ref{experiment} presents datasets, implementation, and experimental results. Section~\ref{discussion} provides a comprehensive discussion and Section~\ref{conclusion} summarizes the study.
\begin{figure*}
     \centering
\includegraphics[width=1\linewidth]{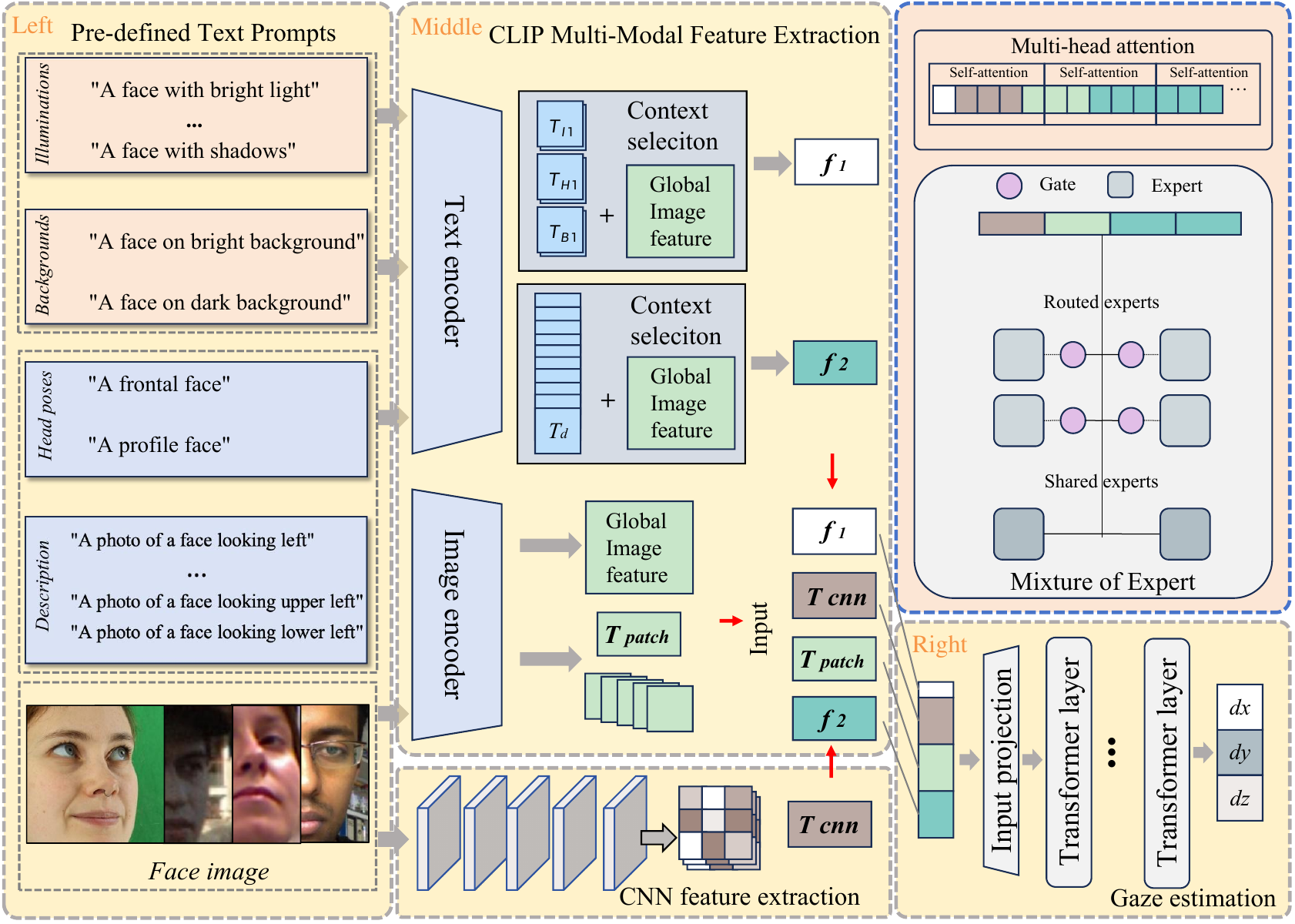}

\caption{Flowchart of GMGaze. \textbf{Left}: Pre-defined prompt banks (Illuminations, Backgrounds, Head poses, and Descriptions) are encoded by the frozen CLIP text encoder to initialize learnable semantic prototype banks. \textbf{Middle}: A CLIP image encoder outputs a global embedding and patch tokens; a CNN provides high-resolution local tokens. The semantic tokens \textit{$\mathbf{\textit{T}}_{I1}$}, \textit{$\mathbf{\textit{T}}_{H1}$}, \textit{$\mathbf{\textit{T}}_{B1}$}, and \textit{$\mathbf{\textit{T}}_{d}$} encode key contextual dimensions: \textit{Illumination}, \textit{Head pose}, \textit{Background}, and \textit{Description}. Prototype similarity and discrete selection produce two conditioned global tokens $f_1$ and $f_2$, which are assembled with patch ($T_{\text{patch}}$) and CNN ($T_{\text{cnn}}$) tokens. \textbf{Right}: All tokens are projected to a unified dimension, concatenated, and processed by stacked transformer layers whose Feed-Forward Networks (FFN) are replaced by routed-and-shared MoE experts to predict the 3D gaze direction $(d_x,d_y,d_z)$.}
    \label{pipeline_main}
\end{figure*}

\section{Related work}\label{RW}



Many studies provide comprehensive overviews of DL based gaze estimation, including single-user and multi-user settings, and have discussed model architectures, datasets, and remaining challenges \citep{pathirana2022eye,cheng2024appearance}. In addition to pure regression models, in \citep{xu2014mimicking}, they propose fusing top-down priors and low-level features to build robust vision models. Furthermore, in \citep{xu2019bio}, they introduce bio-inspired approaches to estimate gaze positions in complex dynamic scenes such as driving, emphasizing low-cost hardware implementation. In this study, we focus on appearance-based single-user 3D gaze estimation and organize related work according to their primary architectural frameworks.

\mypar{CNN based approaches.} In \cite{zhang2017mpiigaze}, they propose the first deep appearance-based gaze estimation method with a 16-layer VGG deep convolutional neural network (GazeNet). The experimental results show that the proposed model outperforms SOTA baselines with large margins (e.g., 22\%). Similarly to GazeNet, in \cite{zhang2017s}, they use a CNN to extract features based on full-face images and then average the feature maps for gaze estimation. The experimental results show that it yields a lower error rate on both 2D and 3D benchmarks. In \cite{cheng2020coarse}, they explore a CNN to extract both coarse-grained features from the face image and fine-grained features from eye images to predict gaze direction (CA-Net). Experiments show that CA-Net achieves the SOTA metric in MPIIGaze and EyeDiap datasets. In \cite{cheng2022puregaze}, they design a self-adversarial framework to purify gaze features (e.g., removing gaze-irrelevant factors such as illumination and identity) while using ResNet-18 to extract them (PureGaze). The experimental results show that it improves all unknown target domains, although it was only trained in the source domain. In \cite{bao2022generalizing}, they augment the face images with different rotation angles, and use those images to generate sub-labels for unsupervised gaze adaptation tasks. This rotation consistency property helps the model to learn robust feature representations. The experimental results demonstrate consistent improvement over the baseline model on four cross-domain gaze estimation tasks. In addition, in \cite{liu2021generalizing}, they propose a plug-and-play gaze adaptation framework (PnP-GA) for generalizable gaze estimation. It introduces a collaborative learning strategy guided by outliers to effectively learn domain-invariant features. Similarly to PnP-GA, in \cite{tian2025disengage}, they propose a Personalized Causal Network (PCNet) for generalizable gaze estimation. It presents a ``Disengage AND Integrate'' perspective for handling personalized information, leveraging causal intervention to capture subject-invariant features and an episodic prototype-based task to extract subject-specific features. Experiments on both in-domain and cross-domain settings demonstrate the usefulness of PCNet. However, the computation overhead of PCNet is considerable, which limits its deployment in real-world applications. In \cite{wang2025test}, they explore the usefulness of ResNet-18 in a test-time gaze adaptation setting. Specifically, the model learns a negligible number of parameters from the prompts. This allows it to adjust the feature extracted from the final layer of the gaze encoder without perturbing the original network. In \citep{khan2024deep}, they propose a CNN-based facial profiling system that jointly estimates head pose, gaze, and eye state in a unified framework for facial and gaze comprehension. However, CNNs capture local features, which may not be sufficient to model subtle context-dependent cues necessary for accurate gaze estimation. 

\mypar{Transformer based approaches.} In \cite{cheng2022gaze}, they first introduce transformers (i.e., pure transformers and hybrid transformers) into gaze estimation tasks (GazeTr-Pure). The experimental results show that transformers can provide competitive performance compared to CNN models. In addition, in \cite{lai2024eye}, they design a transformer-based model for egocentric gaze estimation. It has a novel global-local correlation module to explicitly model the vital connection between the global scene context and local visual information to localize gaze fixation. In \cite{Cheng_2023_ICCV}, they propose a Dual-View Gaze Estimation Network (DV-Gaze) to address the limitations of single-camera gaze estimation by using a pair of images from dual cameras. Specifically, it employs a Dual-View Transformer that encodes camera poses to leverage the geometric relationship between the two gaze directions. However, transformers require more computational resources and training data compared to traditional CNN-based models.

\mypar{CLIP based approaches.} In \cite{wang2023gazeclip}, they propose a novel text-guided gaze estimation framework based on the CLIP model. Furthermore, it introduces a cross attention condenser designed to recalibrate visual and text representations. In \cite{zhang2025clip}, they propose a novel differential contrast training strategy to enhance gaze estimation performance using a pre-trained CLIP (CLIP-DFENet). This method establishes a novel connection between languages and gaze differences of image pairs, using a network with a visual appearance-aware branch and a semantic difference-aware branch for robust feature extraction. In \cite{zhu2025cr}, they convert gaze labels into textual descriptions and use a CLIP to align the features between images and texts with gaze cues. Furthermore, they introduce a regression loss function based on image-text similarity to fine-tune the model. However, the global semantic features can not fully represent fine-grained spatial and angular relationships for gaze estimation.

Unlike previous studies, this study proposes a novel hybrid framework for gaze estimation consisting of a CNN, a CLIP, and a transformer. In addition, it introduces an MoE module to enable the model to learn the dominant features. Furthermore, we introduce an adversarial adaptation method with a feature separation loss to improve the performance on cross-domain setting.

\section{Methodology}
\label{methodology}
We present GMGaze, a multimodal transformer that combines CLIP-driven semantic context, patch-level tokens, and CNN features to estimate 3D gaze directions. Figure \ref{pipeline_main} illustrates the general pipeline of the proposed model. The critical path for gaze estimation starts from the input image, which is processed in parallel by three branches: (1) a frozen CLIP vision encoder to extract global semantic features and patch tokens, (2) a semantic prototype bank to decompose global features into context-aware components ($f_1, f_2$) via similarity-based conditioning, and (3) a ResNet backbone to extract high-resolution local CNN tokens. These multi-scale tokens are projected to a unified dimension, concatenated into a sequence, and then processed by a series of transformer layers. Each transformer layer employs a routed-and-shared Mixture-of-Experts (MoE) block to dynamically allocate computation based on token-level patterns, before a final regression head predicts the 3D gaze vector $(d_x, d_y, d_z)$.


\subsection{Problem formulation}\label{sec:problem-formulation}

Given a face image \(I \in \mathbb{R}^{H \times W \times 3}\), the task is to predict a 3D unit gaze vector \(\mathbf{g} \in \mathbb{R}^3\). We consider two complementary scenarios: \textit{within-domain} and \textit{cross-domain} evaluation.

\textit{Within-domain.} Let \(\mathcal{E}_i\) be a pre-trained CLIP vision encoder and \(\mathcal{F}_{\text{cnn}}\) be a trainable CNN (ResNet 50) backbone. From image $I$, three types of features are extracted:
\begin{equation}
\underbrace{\begin{bmatrix} 
\mathcal{E}_i(I) \to \boldsymbol{f}_{\mathrm{\textit{global}}} \in \mathbb{R}^{D} & \mathcal{E}_i(I) \to \boldsymbol{T}_{\mathrm{\textit{patch}}} \in \mathbb{R}^{N \times D_p} \\
\mathcal{F}_{\mathrm{cnn}}(I) \to \boldsymbol{T}_{\mathrm{cnn}} \in \mathbb{R}^{M \times C} & - 
\end{bmatrix}}_{\text{Multi-scale feature extraction from image $I$}}
\end{equation}
where \(\boldsymbol{f}_{\mathrm{\textit{global}}}\) is the CLIP global embedding with a dimension of \(D\) (e.g., $D=512$ for ViT-B/32), \(\boldsymbol{T}_{\mathrm{\textit{patch}}}\) are \(N\) CLIP patch tokens of dimension \(D_p\) (e.g., $D_p=768$), and \(\boldsymbol{T}_{\mathrm{\textit{cnn}}}\) are \(M=H'W'\) tokens obtained by flattening the CNN feature map \(\boldsymbol{F}_{\mathrm{\textit{cnn}}} \in \mathbb{R}^{H'\times W'\times C}\). $H'$ and $W'$ are height and width, respectively.

\textit{Cross-domain.} We are given a labeled source domain $\mathcal{D}_s = \{(I_s^i, \mathbf{g}_s^i)\}_{i=1}^{N_s}$ and an unlabeled target domain $\mathcal{D}_t = \{I_t^j\}_{j=1}^{N_t}$, where the data distributions $\mathcal{D}_s$ and $\mathcal{D}_t$ differ. The goal is to train a model on $\mathcal{D}_s$ that performs well on the target domain $\mathcal{D}_t$. Our approach first aligns the domains via adversarial training and then refines the model using a small labeled subset (100 samples) of the target domain, which is obtained by shuffling the target set indices once with a fixed random seed and taking the first 100 as the supervised subset.

\subsection{Context-aware semantic conditioning}
The standard global embedding $\boldsymbol{f}_{\mathrm{global}}$ from CLIP acts as a monolithic vector that entangles diverse semantic factors such as gaze-relevant cues (e.g., head pose) and nuisance variables (e.g., background clutter, extreme illumination), see Table .\ref{prompt}. Using such an entangled representation directly for regression introduces negative correlations, where the model may overfit dominant, high-variance factors (e.g., lighting intensity) while ignoring subtle variations in the morphology of the eye region. This entanglement reduces the ability of the regression model to identify predictive gaze factors from environmental noise. To resolve this issue, we propose a semantic conditioning mechanism based on learnable prototype banks. Theoretically, this approach aims to project the entangled $\boldsymbol{f}_{\mathrm{global}}$ into factor-specific subspaces. By explicitly conditioning the global representation on distinct contextual priors (e.g., illumination vs. head pose), we reduce the mutual interference between conflicting semantic attributes. This acts as a soft disentanglement constraint, ensuring that $\boldsymbol{f}_1$ and $\boldsymbol{f}_2$ provide complementary, low redundancy information to subsequent MoE layers. This allows experts to focus on task-relevant features instead of processing noisy mixed-distribution inputs.
Specifically, we consider four distinct learnable banks, denoted as $\mathcal{P}_c$, each corresponding to a key context category $c$:  illumination (illum), head pose, background (bg), and description (desc). We designed a fixed vocabulary of descriptive phrases covering common variations in gaze estimation scenarios to formulate these pre-defined text prompts. For instance, the \textit{illum} bank includes prompts such as ``a face with bright light,'' while the \textit{head pose} bank contains ``a frontal face'' Similarly, the \textit{bg} and \textit{desc} banks consist of generic context sentences (e.g., `` face
on dark background''). Table 1 summarizes the text prompts used in this study. Each bank is represented as a matrix of learnable prototype vectors, which are initialized using the feature embeddings of these pre-defined text prompts, encoded by a frozen CLIP text encoder:
\begin{equation}
\small
\mathbf{P} = \begin{bmatrix} 
\mathbf{P}_{\text{illum}} & \mathbf{P}_{\text{head}} \\
\mathbf{P}_{\text{bg}} & \mathbf{P}_{\text{desc}} 
\end{bmatrix},
\quad
\mathbf{P}_{c}\in\mathbb{R}^{D\times K_c},
\end{equation}
where $D$ represents the feature dimension of the output of the CLIP visual encoder, and $K_c$ is the number of prototypes in the bank $c$. The sizes $K_c$ for the four banks are 4, 2, 2 and 8 (i.e., the number of prompt phases defined in Table \ref{tab:prompts}), respectively. Given the normalized global CLIP vector $\hat{\boldsymbol{f}}_{\mathrm{global}}\in\mathbb{R}^{1\times D}$ ($l_2$ normalization), we compute its similarity to all prototypes using a softmax function with a hyperparameter $\tau$:
\begin{equation}\label{eq:context-sel}
\mathbf{S} \;=\; \mathrm{softmax}\!\Big(\tau\cdot \hat{\boldsymbol{f}}_{\mathrm{global}}\hat{\mathbf{P}}^\top\Big)
\;=\;
\begin{bmatrix}
s_{\text{illum}} & s_{\text{head}}\\[2pt]
s_{\text{bg}} & s_{\text{desc}}
\end{bmatrix},
\end{equation}
where $s_c\in\mathbb{R}^{1\times K_c}$ are the similarity scores for bank $c$ and $\hat{\mathbf{P}}$ is the column-normalized matrix of all prototypes. $\tau$ is the temperature coefficient. This hyperparameter serves as a scaling factor to modulate the sharpness of the probability distribution over the available prototypes.

For each class, we employ a discrete selection mechanism to enforce a semantic bottleneck, preventing the model from averaging conflicting priors. To ensure differentiability across this discrete choice, we use the Straight-Through Estimator (STE) \citep{bengio2013estimating}. Despite the selection mechanism can be achieved using a soft-weighted fusion (i.e., expected value) of all prototypes within each bank, however, we argue that the prototypes in the proposed banks are designed to represent coarse and potentially conflicting contextual priors rather than continuous basis functions. The primary goal is to impose a semantic bottleneck and provide a dominant context cue for downstream MoE routing, rather than to reconstruct a continuous context manifold. Averaging multiple prototypes may combine these distinct semantic meanings and introduce noise variables. In contrast, hard selection preserves one dominant prior per bank in the forward pass, while STE ensures this discrete selection remains fully end-to-end trainable.
Specifically, let $\mathbf{e}_c^{\text{hard}}$ be the one-hot vector corresponding to the index of maximum probability in $\mathbf{s}_c$. We formulate the selection vector $\mathbf{e}_c$ used for conditioning as:
\begin{equation}
    \mathbf{e}_c = \mathbf{e}_c^{\text{hard}} - \mathbf{s}_c.\mathrm{detach()} + \mathbf{s}_c
\end{equation}
In the forward pass, the term $(\mathbf{s}_c - \mathbf{s}_c.\mathrm{detach()})$ evaluates to zero, so the output is effectively the hard selection $\mathbf{e}_c^{\text{hard}}$. In the backward pass, gradients propagate through the soft probabilities $\mathbf{s}_c$ since the hard term is detached. This mechanism allows the model to simultaneously update the prototype values (via the additive path) and refine the routing logic (via the similarity metric in Eq.~\ref{eq:context-sel}) to select more task-relevant prototypes over time. The operator $\mathrm{detach}(\cdot)$ returns the same value in the forward pass but treats the variable as a constant during back-propagation. Consequently, the second term $-\mathbf{s}_c.\mathrm{detach()} + \mathbf{s}_c$ is numerically zero in the forward pass, ensuring the output $\mathbf{e}_c$ represents the hard one-hot selection $\mathbf{e}_c^{\text{hard}}$. During back-propagation, gradients flow exclusively through the last $\mathbf{s}_c$ alone without the detached branch.
Specifically, the prompt vocabulary (i.e., actual textual words) and the CLIP text encoder are frozen. Furthermore, the CLIP prompts are used exclusively to extract the initial feature embeddings for the prototype banks $\mathbf{P}$ without token-level prompt learning or text-encoder fine-tuning.  During the subsequent training of the gaze estimation network, only the numerical prototype vectors in $\mathbf{P}$ are learnable parameters. These vectors are updated via backpropagation to align with the specific gaze dataset distributions, without changing the original textual semantics or the parameters within the CLIP backbone.
\begin{equation}\label{eq:context-vectors}\small
\begin{aligned}
\boldsymbol{f}_1 &= \mathrm{LayerNorm}\Big(\boldsymbol{f}_{\mathrm{global}} + 
[\mathbf{P}_{\mathrm{illum}};\mathbf{P}_{\mathrm{bg}}]\;
[\mathbf{e}_{\mathrm{illum}};\mathbf{e}_{\mathrm{bg}}]\Big),\\[4pt]
\boldsymbol{f}_2 &= \mathrm{LayerNorm}\Big(\boldsymbol{f}_{\mathrm{global}} + 
[\mathbf{P}_{\text{desc}};\mathbf{P}_{\mathrm{head}}]\;[\mathbf{e}_{\text{desc}};\mathbf{e}_{\mathrm{head}}]\Big).
\end{aligned}
\end{equation}

For cross-domain setting, this separation is further enforced during training by a feature separation loss (Section~\ref{sep_loss}), which encourages $\boldsymbol{f}_1$ and $\boldsymbol{f}_2$ to be orthogonal, thereby maximizing their informational complementarity

\begin{table}[htbp]
\centering
\caption{The predefined text prompt vocabulary used for prototype initialization.}
\label{tab:prompts}

\begin{tabular}{lp{5.5cm}}
\toprule
\textbf{Category} & \textbf{Text Prompt Phrases} \\
\midrule
\textit{Illumination} & "a face with bright light", "a face with low light", "a face with shadows" \\
\addlinespace
\textit{Head Pose} & "a frontal face", "a profile face" \\
\addlinespace
\textit{Background} & "a face on bright background", "a face on dark background" \\
\addlinespace
\textit{Description} & "A photo of a face looking left", "A photo of a face looking upper left", "A photo of a face looking up", "A photo of a face looking upper right", "A photo of a face looking right", "A photo of a face looking lower right", "A photo of a face looking down", "A photo of a face looking lower left" \\
\bottomrule
\end{tabular}
\label{prompt}
\end{table}
\begin{figure}
     \centering
\includegraphics[width=1\linewidth]{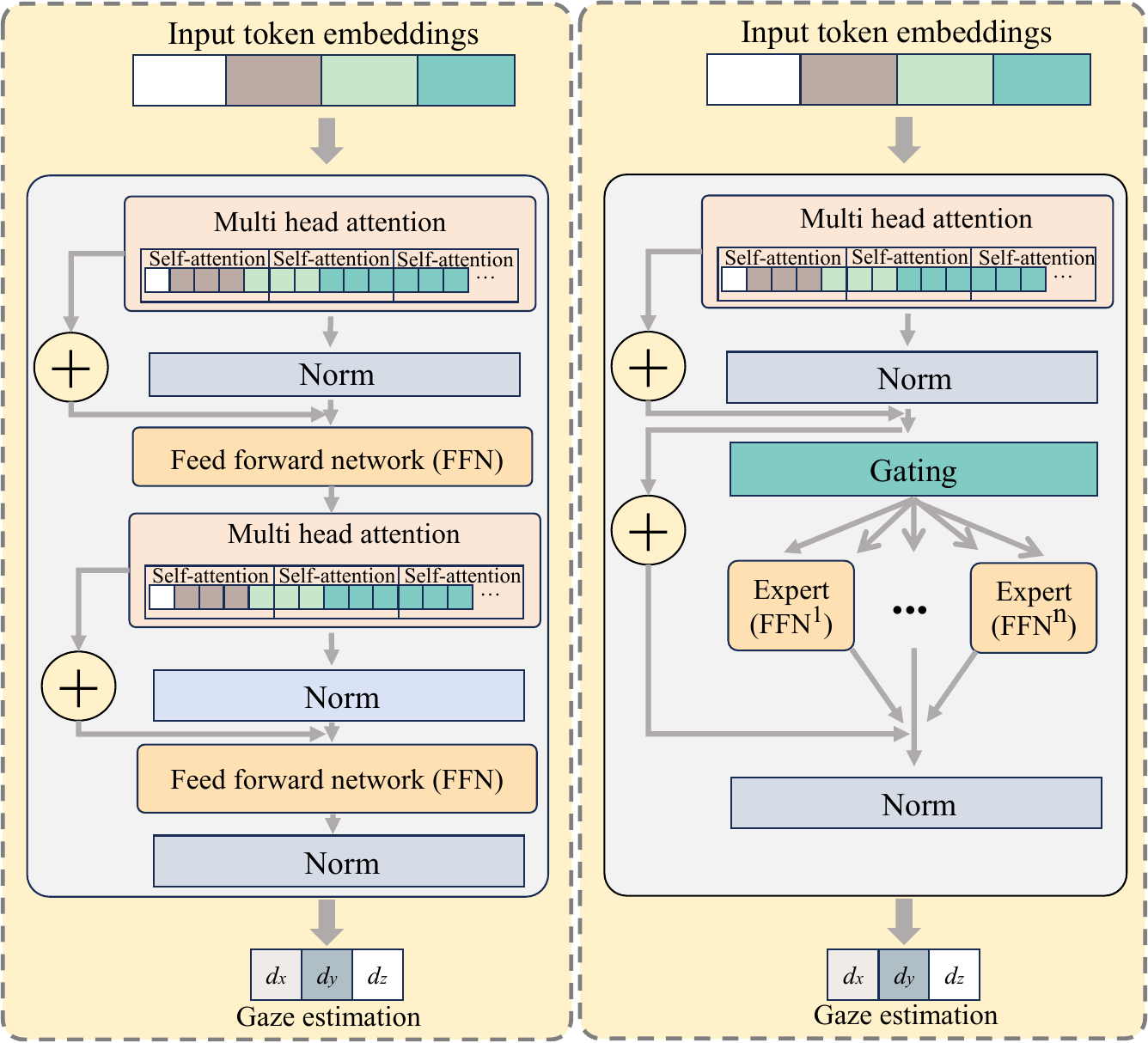}

\caption{\textbf{(Left)}: Example of dense transformer layer, which consists of multi head attention (MHA) followed by a FFN. \textbf{(Right)}: MoE transformer layer in which a FFN block is replaced by a set of experts, which operate in parallel. The final prediction head ultimately yields the 3D gaze direction predictions ($d_x$, $d_y$, $d_z$).}
    \label{moevis}
\end{figure}

\subsection{Unified multi-scale token fusion}

Since these features are extracted from different networks, they have different dimensions. Specifically, the global features are vectors, while the CNN and patch features are sequences with differing channel depths as follows.
\begin{equation}\label{EQ:size}
    \begin{bmatrix}
\boldsymbol{f}_1, \boldsymbol{f}_2 & \in \mathbb{R}^{B \times D} & (B,\,512) \\
\boldsymbol{T}_{\text{cnn}} & \in \mathbb{R}^{B \times M \times C} & (B,\,49,\,2048) \\
\boldsymbol{T}_{\text{patch}} & \in \mathbb{R}^{B \times N \times D_p} & (B,\,49,\,768)
\end{bmatrix}
\end{equation}
where $B$ is the batch size. To fit the input of the Transformer (i.e., the same feature dimension), we project these features into a unified dimension, $d_{\text{model}}$ (e.g., $768$), using separate linear layers:
\begin{equation}\label{eq:projection}
\begin{aligned}
\boldsymbol{f}'_1, \boldsymbol{f}'_2 &= \mathrm{Proj}_{\text{sem}}(\boldsymbol{f}_1), \mathrm{Proj}_{\text{sem}}(\boldsymbol{f}_2) \quad &\in \mathbb{R}^{B \times 1 \times d_{\text{model}}} \\
\boldsymbol{T}'_{\text{patch}} &= \mathrm{Proj}_{\text{patch}}(\boldsymbol{T}_{\text{patch}}) \quad &\in \mathbb{R}^{B \times N \times d_{\text{model}}} \\
\boldsymbol{T}'_{\text{cnn}} &= \mathrm{Proj}_{\text{cnn}}(\boldsymbol{T}_{\text{cnn}}) \quad &\in \mathbb{R}^{B \times M \times d_{\text{model}}}
\end{aligned}
\end{equation}

Note that for $\boldsymbol{f}_1$ and $\boldsymbol{f}_2$, we unsqueeze the explicit sequence dimension to be $1$. Finally, we concatenate these features along the token (sequence) dimension to form the input sequence $\mathcal{X} \in \mathbb{R}^{B \times (2+N+M) \times d_{\text{model}}}$ for the Transformer:
\begin{equation}\label{eq:seq}
\mathcal{X} = \text{Concatenate}\big(\boldsymbol{f}'_1, \boldsymbol{f}'_2, \boldsymbol{T}'_{\text{patch}}, \boldsymbol{T}'_{\text{cnn}}\big)
\end{equation}

Specifically, the proposed fusion mechanism involves three distinct spatial granularities: (1) Global-scale tokens $\boldsymbol{f}_1$ and $\boldsymbol{f}_2$,  derived from the CLIP image-level embeddings, which represent the entire facial context under different semantic conditioning; (2) Intermediate-scale tokens $\boldsymbol{T}_{\text{patch}}$, which retain a coarse spatial grid from CLIP patches and encode region-level semantic structures and (3) Local fine-scale tokens $\boldsymbol{T}_{\text{cnn}}$, extracted from high-resolution CNN feature maps to preserve detailed peri-ocular textures and shading cues.

\subsection{Mixture of Experts Transformer}

In unconstrained gaze estimation, the intrinsic heterogeneity of input samples poses a challenge for a standard dense Transformer, whose position-wise feed-forward network (FFN) applies the same computation to all tokens. Specifically, gaze samples may correspond to markedly different visual conditions, such as frontal well-lit faces, low-illumination scenes, or large head-pose variations. In our setting, the input sequence is also multimodal, as it concatenates prototype-conditioned semantic tokens with patch-level and CNN-based local visual tokens. These token groups reside in different statistical manifolds and may benefit from different computational pathways. To address this issue, we replace the dense FFN with a sparse Mixture-of-Experts (MoE) block, which enables token-wise conditional computation and allows different experts to specialize in different token patterns or environmental conditions. Figure \ref{moevis} shows the differences between dense FFN and MoE.

For clarity, we omit the batch dimension in the following derivation and describe the MoE operation for a single sample. Let the input sequence be
\begin{equation}
\mathcal{X} = [\mathbf{x}_1,\ldots,\mathbf{x}_L]^\top \in \mathbb{R}^{L \times d},
\end{equation}
where $L$ is the sequence length and each token $\mathbf{x}_j \in \mathbb{R}^{d}$ denotes one semantic, patch, or CNN token. Routing is performed independently for each token. For the $j$-th token, the router predicts a distribution over $E$ experts:
\begin{equation}\label{eq:router}
\mathbf{r}_j = \operatorname{softmax}\!\big(\operatorname{Gating}(\operatorname{LayerNorm}(\mathbf{x}_j))\big)\in\mathbb{R}^{E}.
\end{equation}
A Top-$K$ binary mask $\mathbf{m}_j\in\{0,1\}^{E}$ is then applied to retain only the selected experts, yielding sparse routing weights
\begin{equation}
\tilde{\mathbf{r}}_j = \mathbf{r}_j \odot \mathbf{m}_j.
\end{equation}

Each expert is implemented as a position-wise FFN:
\begin{equation}\label{eq:expert}
\operatorname{Expert}_i(\mathbf{x}_j)
=
W_{2,i}\,\sigma(W_{1,i}\mathbf{x}_j + b_{1,i}) + b_{2,i},
\end{equation}
where $W_{1,i}\in\mathbb{R}^{d_{\mathrm{ff}}\times d}$, $W_{2,i}\in\mathbb{R}^{d\times d_{\mathrm{ff}}}$, and $\sigma(\cdot)$ denotes the activation function. The routed MoE output for token $\mathbf{x}_j$ is defined as:
\begin{equation}\label{eq:moe}
\operatorname{MoE}(\mathbf{x}_j)
=
\sum_{i\in\operatorname{TopK}(\mathbf{r}_j)}
\tilde r_{j,i}\,\operatorname{Expert}_i(\mathbf{x}_j).
\end{equation}

Applying this token-wise routing to all tokens yields the sequence output:
\begin{equation}
\operatorname{MoE}(\mathcal{X})
=[\operatorname{MoE}(\mathbf{x}_1),\ldots,\operatorname{MoE}(\mathbf{x}_L)]^\top
\in \mathbb{R}^{L\times d}.
\end{equation}

The MoE block is integrated into each Transformer layer after self-attention. MHA first concatenates information across semantic, patch, and CNN tokens:
\begin{equation}\label{eq:att}
\begin{aligned}
&\operatorname{MultiHead}(\mathcal{X}) =
\operatorname{Concat}_{h=1}^{H}\!\big(A^{(h)}\mathcal{X}W_v^{(h)}\big)W_o,\\
&A^{(h)} =
\operatorname{softmax}\!\Big(
\frac{\mathcal{X}W_q^{(h)}(\mathcal{X}W_k^{(h)})^\top}{\sqrt{d_h}}
\Big)\in\mathbb{R}^{L\times L},
\end{aligned}
\end{equation}
where $H$ is the number of attention heads, $d_h=d/H$ is the per-head dimension, and $W_q^{(h)},W_k^{(h)},W_v^{(h)}\in\mathbb{R}^{d\times d_h}$ and $W_o\in\mathbb{R}^{d\times d}$ are learnable projections. Because $\mathcal{X}$ is organized as modality blocks $(\boldsymbol{f}_1,\boldsymbol{f}_2,\boldsymbol{T}_{\text{patch}},\boldsymbol{T}_{\text{cnn}})$, self-attention naturally enables cross-scale and cross-modal interactions before token-wise expert routing. Each Transformer layer adopts a pre-norm residual form and replaces the standard dense FFN with the MoE block:
\begin{equation}\label{eq:transformer_layer}
\begin{aligned}
\widetilde{\mathcal{X}} &= \mathcal{X} + \operatorname{MultiHead}\big(\operatorname{LayerNorm}(\mathcal{X})\big),\\[4pt]
\mathcal{X} &= \widetilde{\mathcal{X}} + \operatorname{MoE}\big(\operatorname{LayerNorm}(\widetilde{\mathcal{X}})\big),
\end{aligned}
\end{equation}
where $\widetilde{\mathcal{X}}$ indicates the intermediate representation after self-attention. Replacing the dense FFN with MoE increases conditional capacity while preserving residual connectivity and normalization stability. To make modality-aware specialization explicit, we optionally constrain routing by a learnable modality-to-expert mask $\mathbf{S}_{\mathrm{mod}}\in\{0,1\}^{\mathrm{mods}\times E}$ so that tokens from a given modality preferentially activate a subset of experts:
\begin{equation}
\tilde{\mathbf{r}}_j \leftarrow \tilde{\mathbf{r}}_j \odot \mathbf{S}_{\mathrm{mod}(j)},
\end{equation}
where $\mathrm{mod}(j)\in\{\boldsymbol{f}_1,\boldsymbol{f}_2,\boldsymbol{T}_{\mathrm{patch}},\boldsymbol{T}_{\mathrm{cnn}}\}$ identifies the modality of token $\mathbf{x}_j$. This routed-and-shared design allows certain experts to specialize in modality-specific patterns while others remain shared across modalities.

Furthermore, to improve routing and encourage more balanced expert usage, we employ Expert Dropout during training. Specifically, before applying the Top-$K$ mask $\mathbf{m}_j$, we randomly remove a subset of experts with probability $p_{\mathrm{drop}}$. This regularization helps the router to explore alternative expert assignments and decreases over-fitting on a small subset of dominant experts. Overall, the self-attention layers provide dense cross-token fusion, while the MoE layers provide sparse token-dependent capacity allocation. Their combination ensures high-capacity and modality-aware processing within the unified multi-scale token sequence.


\subsection{Prediction head and angular loss.} The Transformer output is pooled and mapped to a 3D vector \(\hat{\mathbf{g}}\). We train with an angular objective that is numerically stable and avoids explicit arccos computation \cite{zhang2017mpiigaze}:
\begin{equation}\label{eq:ang-loss}
\mathcal{L}_{\text{ang}}(\hat{\mathbf{g}},\mathbf{g}) = 1 - \frac{\hat{\mathbf{g}}^\top \mathbf{g}}{\|\hat{\mathbf{g}}\|\|\mathbf{g}\|},
\end{equation}
where $\mathbf{g}$ is the ground truth.

\begin{figure}
     \centering
\includegraphics[width=1\linewidth]{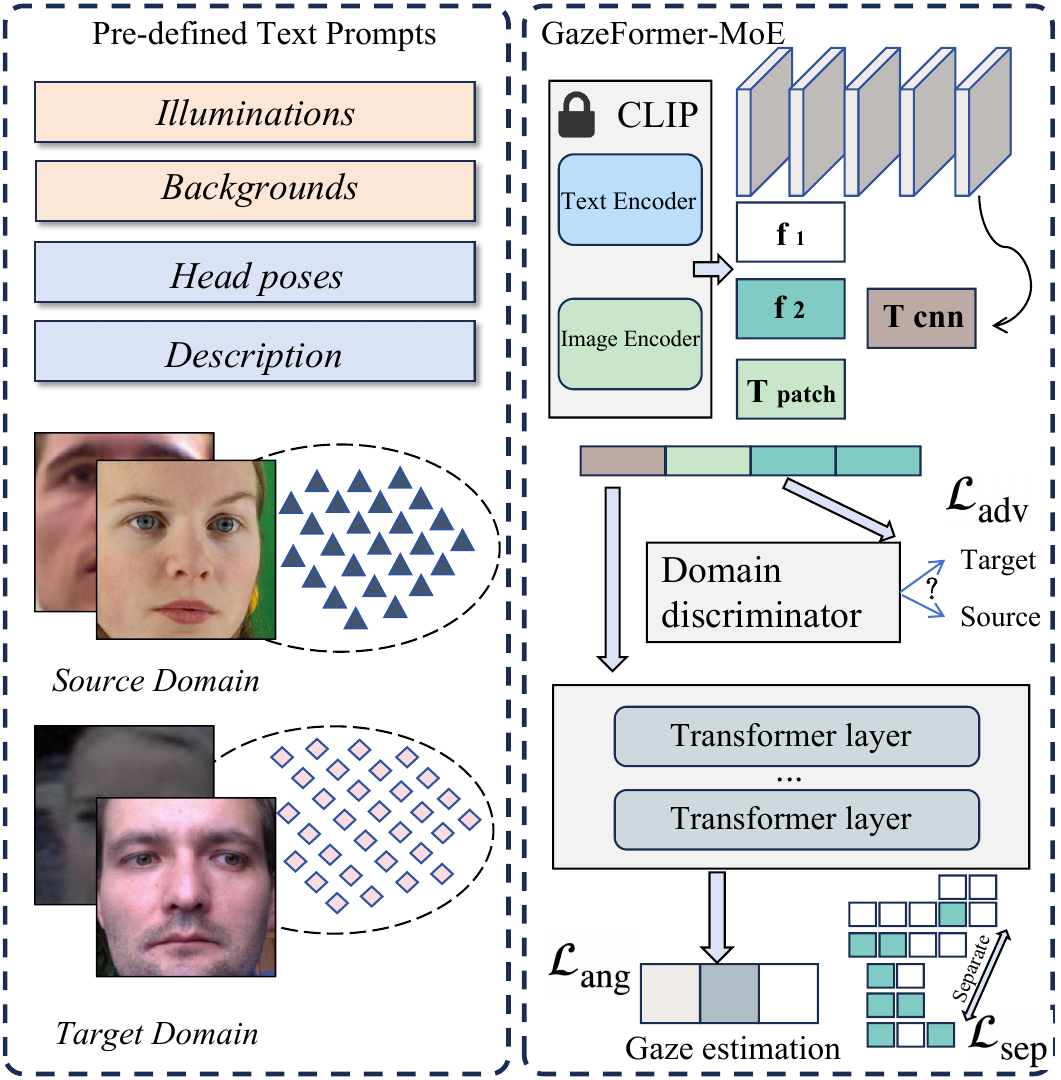}

\caption{
Overall training paradigm of cross-domain setting. During training, a domain discriminator is jointly optimized with the GMGaze to perform adversarial domain adaptation, reducing the distribution discrepancy between source and target domains. Meanwhile, the absolute cosine similarity between the two global semantic feature vectors ($\boldsymbol{f}_1$ and $\boldsymbol{f}_2$) is minimized to separate them into orthogonal directions in the feature space, encouraging the model to learn more gaze-related and domain-invariant representations. Finally, the model yields the 3D gaze direction prediction ($d_x$, $d_y$, $d_z$).}

    \label{sep_da}
\end{figure}

\subsection{Feature adaptation}

\subsubsection{Feature separation loss}
\label{sep_loss}
To separate $\boldsymbol{f}_1$ and $\boldsymbol{f}_2$ and encourage complementary information, we introduce a feature separation loss:
\begin{equation}
\label{eq:sep_loss}
\mathcal{L}_{\text{sep}} = \frac{1}{B} \sum_{i=1}^{B} \left| (\hat{\boldsymbol{f}}_1^i)^\top \hat{\boldsymbol{f}}_2^i \right|,
\end{equation}
where $\hat{\boldsymbol{f}}_{1|2}^i = \boldsymbol{f}_{1|2}^i / \|\boldsymbol{f}_{1|2}^i\|_2$ is the normalized feature vector.

Minimizing $\mathcal{L}_{\text{sep}}$ penalizes the cosine similarity between the two normalized feature representations, thereby pushing $\hat{\boldsymbol{f}}_1$ and $\hat{\boldsymbol{f}}_2$ to orthogonal directions in the feature space. Geometrically, this acts as a soft orthogonality constraint, decorrelating the two branches by reducing feature redundancy and encouraging each branch to capture a distinct, complementary semantic subspace. This separation enhances the diversity of learned features and prevents both branches from encoding overlapping cues.

\subsubsection{Adversarial domain adaptation}

To align the feature distributions, we introduce a domain discriminator $\mathcal{D}$. The discriminator is a multi-layer perceptron (MLP) that takes the concatenated feature vectors $\mathcal{X}^{s|t} = \text{Concatenate}\big(\boldsymbol{f}_1, \boldsymbol{f}_2, \boldsymbol{T}_{\text{patch}},\boldsymbol{T}_{\text{cnn}}\big)$ as input and predicts a domain label (e.g., 0 for source, 1 for target). The discriminator $\mathcal{D}$ is trained to minimize the expectation $\mathbb{E}$ of:

\begin{equation}
\label{eq:adv_loss_D}
\mathcal{L}_{\mathcal{D}} = -\mathbb{E}_{I_s \sim \mathcal{D}_s}[\log \mathcal{D}(\mathcal{X}^s)] - \mathbb{E}_{I_t \sim \mathcal{D}_t}[\log(1 - \mathcal{D}(\mathcal{X}^t))]
\end{equation}

Simultaneously, the GMGaze $\mathcal{G}$ is trained with the following objective:
\begin{equation}
\label{eq:adv_loss_G}
\mathcal{L}_{\text{adv}} = -\mathbb{E}_{I_t \sim \mathcal{D}_t}[\log \mathcal{D}(\mathcal{X}^t)].
\end{equation}
where $\mathbb{E}$ denotes the expectation, i.e., the average value computed over samples drawn from the corresponding source or target domain.

Following \cite{ganin2016domain}, we implement this using a Gradient Reversal Layer (GRL). Figure \ref{sep_da} shows the procedure of the domain adversarial training and the proposed feature separation loss.

\begin{algorithm}[ht]
\caption{Process within-domain training of GMGaze.}
\label{alg:pipeline_simplified}
\small
\begin{algorithmic}
\Require Image $I$, ground-truth gaze $\mathbf{g}$
\Ensure Predicted gaze $\hat{\mathbf{g}}$
\State Extract CLIP global + patch tokens $(\boldsymbol{f}_{\mathrm{global}},\boldsymbol{T}_{\mathrm{patch}})=\mathcal{E}_i(I)$ and CNN tokens $\boldsymbol{T}_{\mathrm{cnn}}=\mathrm{Flatten}(\mathcal{F}_{\mathrm{cnn}}(I))$
\State Compute prototype similarities $\mathbf{S}$ (Eq.~\ref{eq:context-sel}); select $p_c=\arg\max s_c$ (hard routing; stop-gradient through selection)
\State Form context-biased vectors $\boldsymbol{f}_1,\boldsymbol{f}_2$ (Eq.~\ref{eq:context-vectors})
\State Concatenate sequence $\mathcal{X}$ using these features (Eq.~\ref{eq:projection}, ~\ref{eq:seq}.)
\For{$\ell=1$ to $L$}
  \State $\widetilde{\mathcal{X}}=\mathcal{X}+\text{MultiHead}(\mathrm{LayerNorm}(\mathcal{X}))$ (Eq.~\ref{eq:att})
  \State $\mathcal{X}=\widetilde{\mathcal{X}}+\mathrm{LayerNorm}(\text{MoE}(\widetilde{\mathcal{X}}))$ (Eq.~\ref{eq:transformer_layer}, \ref{eq:moe})
\EndFor
\State Compute angular loss $\mathcal{L}_{\text{ang}}$ (Eq.~\ref{eq:ang-loss}); set $\mathcal{L}_{\text{total}}=\mathcal{L}_{\text{ang}}$ (Eq.~\ref{eq:total_loss})
\State \Return $\hat{\mathbf{g}}$
\end{algorithmic}
\end{algorithm}

\subsection{Overall training objective}

\mypar{Within-domain setting.} The final training loss uses angular loss only:
\begin{equation}
\label{eq:total_loss}
\mathcal{L}_{\text{total}} = \mathcal{L}_{\text{ang}}
\end{equation}

Algorithm \ref{alg:pipeline_simplified} illustrates the within-domain training procedure for the GMGaze model.

\begin{algorithm}[ht]
\caption{Process of cross-domain training for GMGaze.}
\label{alg:cross_domain_compact}
\small
\begin{algorithmic}
\Require Labeled source $\mathcal{D}_s$, unlabeled target $\mathcal{D}_t$, model $\mathcal{G}$, discriminator $\mathcal{D}$
\For{each step}
  \State Sample $(I_s,\mathbf{g}_s)\sim\mathcal{D}_s$, $I_t\sim\mathcal{D}_t$
  \State Obtain $\boldsymbol{f}_1, \boldsymbol{f}_2, \boldsymbol{T}_{\text{patch}}, \boldsymbol{T}_{\text{cnn}}$
  for both domains (Eq.~\ref{eq:context-sel}, \ref{eq:context-vectors})
  \State Project and build sequences $\mathcal{X}^s,\mathcal{X}^t$ (Eq.~\ref{eq:projection}, \ref{eq:seq})
  \State Compute $\mathcal{L}_{\text{ang}}$ using $\mathcal{X}^s$ (Eq.~\ref{eq:ang-loss})
  \State Compute separation loss $\mathcal{L}_{\text{sep}}$ (Eq.~\ref{eq:sep_loss})
  \State Compute $\mathcal{L}_{\mathcal{D}}$ (Eq.~\ref{eq:adv_loss_D}) and $\mathcal{L}_{\text{adv}}$ (Eq.~\ref{eq:adv_loss_G})
  \State Update discriminator minimizing $\mathcal{L}_{\mathcal{D}}$ 
  \State Update $\mathcal{G}$ minimizing $\mathcal{L}_{\text{total}}=\mathcal{L}_{\text{ang}}+\mathcal{L}_{\text{adv}}+\mathcal{L}_{\text{sep}}$ (Eq.~\ref{eq:total_loss_2})
\EndFor
\end{algorithmic}
\end{algorithm}

\mypar{Cross-domain setting.} The final training loss combines all components:
\begin{equation}
\label{eq:total_loss_2}
\mathcal{L}_{\text{total}} = \mathcal{L}_{\text{ang}} + \mathcal{L}_{\text{adv}} + \mathcal{L}_{\text{sep}}
\end{equation}

Algorithm \ref{alg:cross_domain_compact} illustrates the cross-domain training procedure for the GMGaze model.


\section{Experiment}
\label{experiment}

\subsection{Datasets}
We evaluate GMGaze on four widely used  benchmarks, including the within-domain evaluation and cross-domain evaluation. 

\mypar{EYEDIAP (\textbf{E, $\mathcal{D}_E$}).} It derives from the video recordings of 14 subjects under the screen target session, one image was sampled every 15 frames—this process ultimately yielded a total of 50022 facial images \cite{funesmora2014eyediap}. 

\mypar{ETH-XGaze (\textbf{Et, $\mathcal{D}_{Et}$}).} In ETH-XGaze, 756,540 images sourced from 80 subjects were used. The dataset itself comprises over one million high-resolution images, encompassing diverse gaze directions, extreme head poses, and complex illumination conditions \cite{zhang2020ethxgaze}.

\mypar{Gaze360 (\textbf{G}, $\mathcal{D}_{G}$).} Gaze360 includes indoor and outdoor environments, which contains 3D gaze information spanning a broad range of head poses and distances—with a total of 238 subjects involved in the data collection process \cite{kellnhofer2019gaze360}.

\mypar{MPIIFaceGaze (\textbf{M, $\mathcal{D}_{M}$}).} The MPIIFaceGaze dataset was collected in everyday life scenarios. It includes 15 participants, and each participant has 3000 image samples associated with them \cite{zhang2017fullface}.

\begin{figure*}
     \centering
\includegraphics[width=0.99\linewidth]{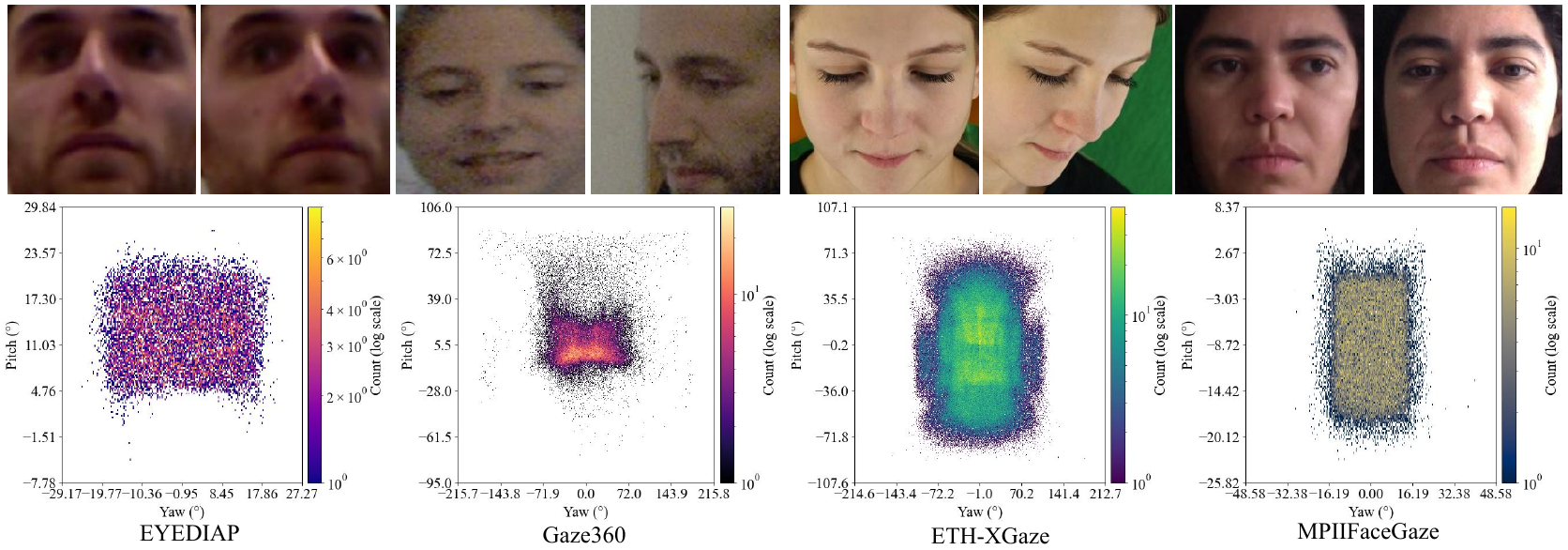}

\caption{Example for samples from the four benchmarks (EYEDIAP, Gaze360, ETH-XGaze, MPIIFaceGaze), showing diversity in illumination, head pose, background, and appearance (\textbf{First row}). 2D gaze direction distributions (yaw vs. pitch, degrees) for these datasets (\textbf{Second row}). Each point corresponds to a labeled gaze sample; the colormap encodes log-scaled sample density (warmer colors = higher frequency).}
    \label{data_distribution}
\end{figure*}

These datasets collectively cover a broad spectrum of conditions, including controlled indoor environments and unconstrained real-world scenes, thus facilitating the evaluation of various lighting, background and head position variations. As illustrated in Figure~\ref{data_distribution}, the gaze angle distributions differ notably among datasets: some are centered around near-frontal views, whereas others present a much wider range of yaw and pitch angles. For Gaze360, we use the official training and testing sets. We follow the protocols in \citep{cheng2024appearance} for MPIIFaceGaze, EYEDIAP and ETH-XGaze.

\subsection{Implementation details}
\label{implementationdetails}

The experiments were performed on NVIDIA RTX 4090 with PyTorch 2.4.1+cu124. The model uses a pre-trained CLIP \cite{radford2021learning} (with a frozen ViT-B/32 image encoder), a pre-trained ResNet-50 backbone, and a 12-layer MoE Transformer. The Transformer has a unified feature dimension ($d_{\text{model}}$) of 768, eight attention heads, and FFNs with a hidden size of 2048. The MoE module consists of four experts, with a Top-2 gating mechanism (two active experts per token). The two experts with the highest gating scores are selected for each token, while the other two can be considered shared, as they are available to all tokens. 
During training, we apply an Expert Dropout rate of 0.2 to encourage load balancing across the experts. Training uses the AdamW optimizer \cite{loshchilov2017decoupled} with a learning rate scheduled by cosine annealing from $10^{-4}$ to $10^{-6}$ over 100 epochs. For with-in and cross-domain evaluations, we use a batch size of 128 and 64, respectively. For the proposed discriminator $\mathcal{D}$, all sequence-based features are first converted into fixed-length vectors via token-wise mean pooling. The resulting vectors are concatenated and projected by an MLP into a 960-dimensional latent space, followed by a two-layer classification head (linear layers with ReLU activation) to predict source or target domains, with a dropout rate of 0.3. Following the study in~\cite{cheng2024appearance}, we report the angular error based on a single run with a fixed random seed, which is empirically set to 0. The CLIP vision and text encoder are kept frozen during training, while we optimize the CNN backbone, the Transformer (self-attention, MoE router, and experts), the projection layers, and the learnable semantic prototype banks $\mathbf{P}$. Since prototype conditioning is implemented as \emph{hard} selection in the forward pass (i.e., choosing the most similar prototype via $\arg\max$), we introduce a Straight-Through Estimator (STE) to achieve back-propagation.

\begin{table}[h]\scriptsize
\centering
\renewcommand{\arraystretch}{1.2}
\caption{Comparison of GE error (in degrees) on multiple datasets.}
\label{tab:gaze_comparison}
\setlength{\tabcolsep}{4pt}
\begin{tabular}{lccccc}
\toprule
\textbf{Methods} & \textbf{Pub. Year} & \textbf{Et} & \textbf{G} & \textbf{E} & \textbf{M} \\
\midrule
Gazenet \cite{zhang2017mpiigaze}          & TPAMI 17     & - & - & 6.79$^\circ$  & 5.76$^\circ$  \\
FullFace \cite{zhang2017s}          & CVPR 17     & 7.38$^\circ$ & 14.99$^\circ$ & 6.53$^\circ$ & 4.93$^\circ$        \\
Dilated-Net \cite{chen2018appearance}          & ACCV 19     & - & 13.73$^\circ$ & 6.19$^\circ$ & 4.42$^\circ$        \\
Gaze360 \cite{kellnhofer2019gaze360}          & ICCV 19     & 11.04$^\circ$ & 11.04$^\circ$ & 5.36$^\circ$ & 4.06$^\circ$        \\
CA-Net \cite{cheng2020coarse}          & AAAI 20     & - & 11.20$^\circ$ & 5.27$^\circ$ & 4.27$^\circ$        \\
AFF-Net \cite{bao2021adaptive}                 & PR 2020     & - & - & 6.41$^\circ$ & 4.92$^\circ$        \\
GazeTR-Hybrid \cite{cheng2022gaze}          & ICPR 22     & - & 11.46$^\circ$ & 5.44$^\circ$ & 4.00$^\circ$        \\
GazeTr-Pure \cite{cheng2022gaze}          & ICPR 22     & - & 13.58$^\circ$ & 5.72$^\circ$ & 4.24$^\circ$        \\
GazeCLIP \cite{wang2023gazeclip}       & arXiv 23    & - & - & 4.70$^\circ$ & 3.50$^\circ$        \\
GTiT -Hybrid \cite{wu2023attention}       & EAAI 23    & - & 11.20$^\circ$ & 5.35$^\circ$ & 4.11$^\circ$        \\

CLIP-DFENet \cite{zhang2025differential}    & ICM 25    & - & 10.54$^\circ$ & 4.97$^\circ$ & 3.71$^\circ$        \\
SAZE \cite{kim2024appearance}    & PR 24    & - & - & 4.42$^\circ$ & 3.89$^\circ$        \\
EG-Net \cite{wu2024eg}    & ESWA 24    & - & - & 4.55$^\circ$ & 2.76$^\circ$        \\
RAGE-net \cite{kuric2025democratizing}       & EAAI 24    & - & - & - & 4.08$^\circ$        \\
PaCo \cite{xia2024joint}    & CVIU 24    & - & 10.27$^\circ$ &  & 3.23$^\circ$        \\
ADGaze \cite{li2025adgaze}    & PR 25    & - & 10.63$^\circ$ & 4.67$^\circ$ & 3.62$^\circ$        \\
ADGazeS \cite{li2025adgaze}    & PR 25    & - & 10.63$^\circ$ & 4.83$^\circ$ & 3.83$^\circ$        \\
MCA-PGI \cite{li2025nonlinear}    & S. Reports 25    & - & 10.34$^\circ$ & 4.58$^\circ$ & 3.90$^\circ$        \\
FSIGaze \cite{jia2025frequency}       & DISPLAYS 25    & 3.47$^\circ$ & - & 4.94$^\circ$ & 3.47$^\circ$        \\
SLYKLatent \cite{adebayo2025slyklatent}    & THMS 25    & 3.77$^\circ$ & 10.70$^\circ$ & - & 3.77$^\circ$        \\
GazeSymCAT \cite{zhong2025gazesymcat}    & JCDE 25    & 3.28$^\circ$ & - & 5.13$^\circ$ & 4.11$^\circ$        \\
IGTGGaze \cite{nie2025iris}    & TIP 25    & - & 10.92$^\circ$ & 4.56$^\circ$ & 3.60$^\circ$        \\
OmniGaze \cite{qu2025omnigaze}    & NeuIPS 25    &  & \textbf{9.12$^\circ$} & 4.07$^\circ$ &  2.97$^\circ$       \\
PCNet \cite{tian2025disengage}    & TIP 25    & 4.00$^\circ$ & - & 4.50$^\circ$ & 3.99$^\circ$        \\
\rowcolor{gray!15} \textbf{Ours} & -          & \textbf{1.44$^\circ$} & 10.16$^\circ$ & \textbf{3.22$^\circ$} & \textbf{2.49$^\circ$}        \\
\bottomrule
\end{tabular}
\end{table}

\begin{table}[h]
\centering
\scriptsize
\setlength{\tabcolsep}{1pt} 
\caption{Comparison of gaze estimation (GE) error in degrees under cross-domain evaluation. Lower values indicate better performance. The best result is highlighted in bold. \textbf{DG} denotes domain generalization (no target-domain samples used), and \textbf{FT} denotes the number of target-domain samples used for fine-tuning.
Part of the results are taken from \cite{cheng2024appearance}.}
\label{tab:gaze_crossdomain}
\begin{tabular}{lcccccc}
\toprule
\textbf{Methods} & \textbf{Pub } & \textbf{FT} & $\mathcal{D}_{Et} \to \mathcal{D}_M$ & $\mathcal{D}_{Et} \to \mathcal{D}_E$ & $\mathcal{D}_{G} \to \mathcal{D}_M$ & $\mathcal{D}_{G} \to \mathcal{D}_E$ \\
\midrule
FullFace \cite{zhang2017s}   & CVPR 17 & 0 & 11.13$^\circ$ & 14.42$^\circ$ & 12.35$^\circ$ & 30.15$^\circ$ \\
CA-Net \cite{cheng2020coarse}   & AAAI 20 & 0 & - & - & 27.13$^\circ$ & 31.41$^\circ$ \\
PnP-GA \cite{liu2021generalizing}   & ICCV 21 & 100 & 6.00$^\circ$ & 6.17$^\circ$ & 5.74$^\circ$ & 7.04$^\circ$ \\
RAT \cite{bao2022generalizing}      & CVPR 22 & 0 & 7.40$^\circ$ & 6.91$^\circ$ & 7.69$^\circ$ & 7.08$^\circ$ \\
PureGaze \cite{cheng2022puregaze}   & AAAI 22 & 0 & 9.28$^\circ$ & 9.32$^\circ$ & 7.08$^\circ$ & 7.48$^\circ$ \\
RUDA \cite{bao2022generalizing}     & CVPR 22 &100 & 5.78$^\circ$ & \textbf{5.10$^\circ$} & 6.88$^\circ$ & 6.73$^\circ$ \\
CRGA-100 \cite{wang2022contrastive} & CVPR 22 & >0 & 5.68$^\circ$ & 5.72$^\circ$ & 6.09$^\circ$ & 6.68$^\circ$ \\
Jitter-Gaze \cite{liu2022jitter} & Arxiv 22 & 100 & 5.35$^\circ$ & 6.62$^\circ$ & 7.18$^\circ$ & 8.61$^\circ$ \\
CRGA \cite{wang2022contrastive}     & CVPR 22 & >0 & 5.48$^\circ$ & 5.66$^\circ$ & 5.89$^\circ$ & \textbf{6.49$^\circ$} \\
LatentGaze \cite{lee2022latentgaze}
& ACCV 22 & 100 & 5.21$^\circ$ & 7.81$^\circ$ & - & \\

SAZE \cite{kim2024appearance}       & PR 24   & 0 & 5.89$^\circ$ & 10.3$^\circ$ & 6.23$^\circ$ & 7.86$^\circ$ \\
GHR-2D \cite{hu2025ghr}             & EAAI 25 & 0 & 6.59$^\circ$ & 8.05$^\circ$ & - & - \\
PnP-GA+ \cite{liu2024pnp}           & TPAMI 24& 100 & 5.34$^\circ$ & 5.73$^\circ$ & 6.10$^\circ$ & 7.62$^\circ$ \\
PCNet \cite{tian2025disengage}      & TIP 25  & 0 & 6.55$^\circ$ & 6.17$^\circ$ & - & - \\
IGTGGaze \cite{nie2025iris}         & TIP 25  & 0 & 6.31$^\circ$ & 7.28$^\circ$ & 6.96$^\circ$ & 7.11$^\circ$ \\
\rowcolor{gray!15} \textbf{Ours (DG)}    & -         & 0 & 5.23$^\circ$ & 8.71$^\circ$ & 5.97$^\circ$ & 7.38$^\circ$ \\
\rowcolor{gray!15} \textbf{Ours (FT)}    & -         & 100 & \textbf{4.81$^\circ$} & 7.99$^\circ$ & \textbf{5.53$^\circ$} & 6.90$^\circ$ \\
\bottomrule
\end{tabular}
\end{table}

\subsection{Experimental results}
For both the with-in and cross-domain setting, since both the benchmark baselines and previous SOTA methods adopt single-run reporting without mean$\pm$standard deviation, we follow the same setting for a consistent comparison.

\begin{figure*}
     \centering
\includegraphics[width=1\linewidth]{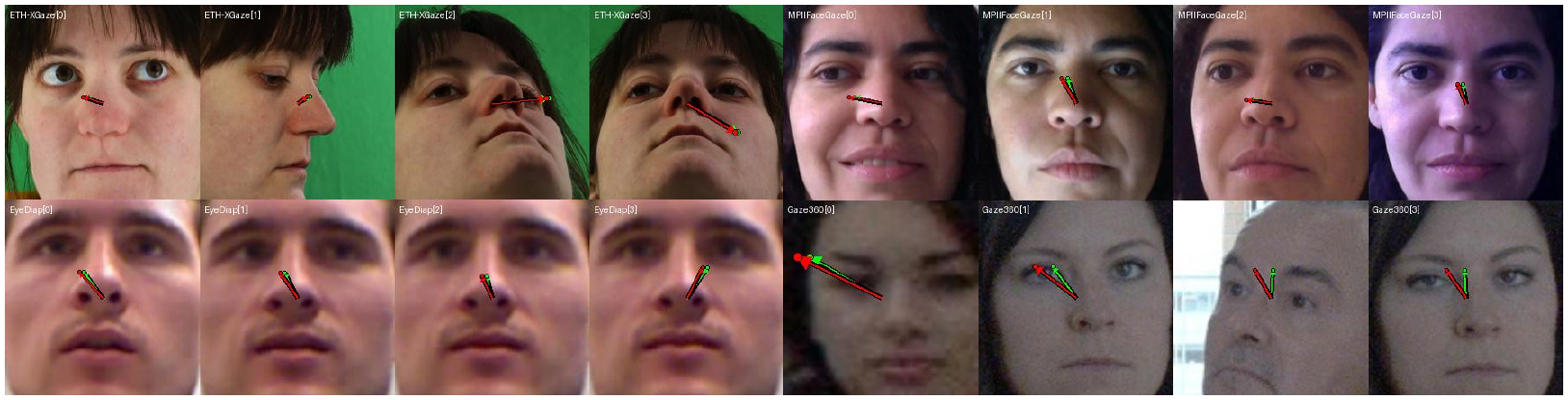}\\
\includegraphics[width=1\linewidth]{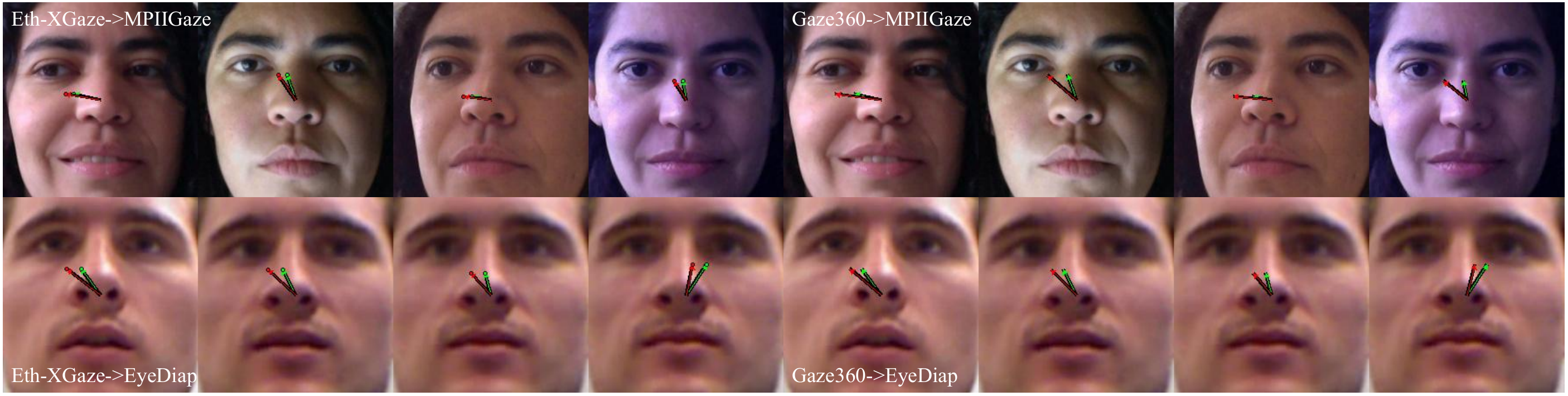}
\caption{Visualization of predictions in within dataset (\textbf{1-2 row}) and cross-domain (\textbf{3-4 row}) evaluation. Green arrows represents the ground truth, red arrows means the predictions.}
    \label{Vis_within}
\end{figure*}

\mypar{Within-domain results.}
As reported in Table~\ref{tab:gaze_comparison}, our method achieves SOTA performance across all four within-domain benchmarks. For example, on the \textbf{M}/\textbf{E} datasets, the proposed model yields an angular error of 2.49$^\circ$/3.22$^\circ$, which is lower than the previous SOTA error of 3.5$^\circ$/4.5$^\circ$. On most challenging dataset \textbf{G}, the proposed model improves the angular error to 10.16$^\circ$ compared to previous SOTA (10.34$^\circ$). Notably, the largest performance increase appears on \textbf{Et} (from 3.28$^\circ$ to 1.44$^\circ$), consistent with our design goal of handling illumination and background diversity: unified multi-scale token fusion with routed and shared MoE markedly mitigates overfitting to studio-like lighting and captures fine periocular shading cues.  Figure~\ref{Vis_within} shows qualitative prediction examples on the four benchmarks (with green representing the ground truth and red indicating predictions). As illustrated, our method exhibits a similar direction compared to the ground truth on \textbf{E}, \textbf{Et} , and \textbf{M}.

\mypar{Cross-domain results.} Table~\ref{tab:gaze_crossdomain} shows that the proposed model achieves the best results on two cross-domain benchmarks, namely $\mathcal{D}_{Et}\!\to\!\mathcal{D}_M$ (4.81$^\circ$, 0.53$^\circ$ lower compared to the previous SOTA) and $\mathcal{D}_{G}\!\to\!\mathcal{D}_M$ (5.53$^\circ$, 0.12$^\circ$ lower than PnP-GA). On $\mathcal{D}_{G}\!\to\!\mathcal{D}_E$, the proposed model achieves a comparable angular error compared to CRGA (6.90$^\circ$ vs. 6.49$^\circ$ CRGA). This suggests that the proposed model is sensitive to the controlled but low diversity of the EYEDIAP backgrounds, whereas the CRGA explicitly regularizes using contrastive augmentation. The largest performance gap is on $\mathcal{D}_{Et}\!\to\!\mathcal{D}_E$ (7.99$^\circ$ vs. 5.10$^\circ$ with RUDA). We attribute this to (i) prototype banks learned on ETH-XGaze over-emphasizing extreme head poses under complex lighting, causing the model to over-specialize in non-dominant features that degrade performance on EYEDIAP; (ii) limited target diversity makes adversarial alignment less stable. Figure~\ref{Vis_within} presents a visualization of cross-domain comparisons, and the results show that the proposed model provides a lower error rate on the target domain after the use of feature separation loss and domain alignment.

\begin{table}[h]
\scriptsize
    \setlength{\tabcolsep}{8.5pt}
    \renewcommand{\arraystretch}{1.1}
    \caption{Impact of fusion strategies on Et, G, E and M datasets.}
    \label{tab:ablation_fusion}
    \begin{tabular}{lcccc}
        \toprule
        \multirow{2}{*}{\textbf{Fusion Strategy}} & \multicolumn{4}{c}{\textbf{Angular Error ($^\circ$)}} \\
        \cmidrule(lr){2-5}
        & \textbf{Et} & \textbf{G} & \textbf{E} & \textbf{M} \\
        \midrule
        Late Fusion          & 1.47$^\circ$ & 11.11$^\circ$ & 4.39$^\circ$ & 3.33$^\circ$ \\
        Intermediate Fusion & 1.58$^\circ$ & 11.10$^\circ$ & 4.34$^\circ$ & 3.60$^\circ$ \\
        \textbf{Early Fusion (Ours)} 
        & \textbf{1.44$^\circ$} 
        & \textbf{10.16$^\circ$} 
        & \textbf{3.22$^\circ$} 
        & \textbf{2.49$^\circ$} \\
        \bottomrule
    \end{tabular}
\end{table}

\begin{table}[ht]
    \centering\scriptsize
    \setlength{\tabcolsep}{6pt}
    \renewcommand{\arraystretch}{1}
    \caption{Ablation study on different feature combinations. $\checkmark$ indicates the feature is used. Best result is highlighted with \textbf{bold} text.}
    \label{tab:ablation1}
    \begin{tabular}{cccccccc}
        \toprule
        \multicolumn{4}{c}{\textbf{Feature combination}} & \multicolumn{4}{c}{\textbf{Datasets}} \\
        \cmidrule(lr){1-4} \cmidrule(lr){5-8}
        $\boldsymbol{f}_1$ & $\boldsymbol{f}_2$ & $\boldsymbol{T}_{\text{cnn}}$ & $\boldsymbol{T}_{\text{patch}}$
        & \textbf{Et} & \textbf{G} & \textbf{E} & \textbf{M} \\
        \midrule
        $\checkmark$ & $\times$ & $\times$ & $\times$ & 10.75$^\circ$ & 28.43$^\circ$ & 10.25$^\circ$ & 7.66$^\circ$ \\
        $\checkmark$ & $\checkmark$ & $\times$ & $\times$ & 10.40$^\circ$ & 27.70$^\circ$ & 10.15$^\circ$ & 7.72$^\circ$ \\
        $\checkmark$ & $\checkmark$ & $\checkmark$ & $\times$ & 1.66$^\circ$ & 10.92$^\circ$ & 4.39$^\circ$ & 3.20$^\circ$ \\
        $\checkmark$ & $\checkmark$ & $\checkmark$ & $\checkmark$ & \textbf{1.44}$^\circ$ & \textbf{10.16}$^\circ$ & \textbf{3.22}$^\circ$ & \textbf{2.49}$^\circ$ \\
        \bottomrule
    \end{tabular}
\end{table}

\begin{table}[htbp]
\centering
\caption{Hardware complexity and efficiency analysis of GMGaze.}
\label{tab:complexity}
\resizebox{1.0\columnwidth}{!}{%
\begin{tabular}{@{}llcc@{}}
\toprule
\textbf{Metric Type} & \textbf{Component / Metric} & \textbf{Value} & \textbf{Notes} \\
\midrule
 & Total Parameters & 165.20 & \textit{Frozen + Trainable} \\
 & Frozen (CLIP Image Encoder) & 86.00 & \textit{Zero training memory overhead} \\
 & Trainable (ResNet, MoE, Protos) & 79.20 & \textit{Updated via Backprop} \\
 \multirow{-4}{*}{Parameters (M)} & \textbf{Active for Inference} & \textbf{129.03} & \textit{Sparse Activation (MoE)} \\
\midrule
 & Frozen CLIP Vision Encoder & 17.50 & \textit{Fixed feature extraction} \\
 & ResNet-50 Backbone & 4.12 & \textit{Local feature extraction} \\
 & MoE Transformer Core & 5.52 & \textit{Top-2 Routing} \\
 \multirow{-4}{*}{FLOPs (G)} & \textbf{Total Inference FLOPs} & \textbf{27.14} & \textit{Full Pipeline} \\
\bottomrule
\end{tabular}%
} 
\end{table}

\subsection{Ablation study}

\mypar{Fusion strategies.}
Table~\ref{tab:ablation_fusion} reports the test metric on three different fusion strategies of the two prototype-conditioned global with visual tokens. Specifically, early fusion concatenates $\{\boldsymbol{f}'_1,\boldsymbol{f}'_2,\boldsymbol{T}'_{\text{patch}},\boldsymbol{T}'_{\text{cnn}}\}$ before the first Transformer layer (Eq.~\ref{eq:seq}). {Intermediate fusion} performs fusion by cross-attention between semantic tokens and visual tokens in intermediate layers. {Late fusion} merges modality representations at the prediction stage. As reported, early fusion consistently provides the best metrics: \textbf{G} improves from 11.10/11.11$^\circ$ to 10.16$^\circ$ ($\sim$0.95$^\circ$), \textbf{E} from 4.34--4.39$^\circ$ to 3.22$^\circ$ ($\sim$1.1$^\circ$), and \textbf{M} from 3.33--3.60$^\circ$ to 2.49$^\circ$ (0.84--1.11$^\circ$). While, {intermediate fusion} (cross-attention) and {late fusion} are less effective due to the constraints on interaction and limited guidance on local peri-ocular cues during encoding.

\mypar{Feature combination.} We validate the contribution of each feature (e.g., $\boldsymbol{f}_1$) on M, E, G and Et datasets. As reported in Table~\ref{tab:ablation1}, using $\boldsymbol{f}_1$ alone fails to capture dominant gaze features and results in a higher error rate of 7.66$^\circ$ on the M dataset. Furthermore, introducing $\boldsymbol{T}_{\text{cnn}}$ considerably improves the performance of the model, resulting in error rates of 3.20$^\circ$ and 10.92$^\circ$ on the M and G datasets, respectively. Finally, combining all features yields the lowest error rate on all datasets (e.g., 10.16$^\circ$ on the G dataset). These results confirm the usefulness of these combined features.

\mypar{Complexity and efficiency analysis.}
Table~\ref{tab:complexity} summarizes the computation overhead of GMGaze. As reported, although GMGaze has a total of 165.20 M parameters, its improved performance is not the result of increasing the training load. Instead, we froze the CLIP image encoder to leverage its robust global semantic priors. This resulted in a feasible trainable parameter of 79.20 M within the CNN backbone and the sparse transformer layers of the MoE. This ensures that the model remains highly adaptable to gaze-specific features without incurring the overhead of a fully dense architecture. For inference cost, although the frozen CLIP and ResNet branches contribute largely to the inference load (e.g., approximately 79.7\% of the total 27.14 GFLOPs), the MoE Transformer introduces a relatively small overhead of 5.52 GFLOPs. Specifically, the Top-2 expert routing mechanism ensures that for any input token, only a sparse subset of the available parameters is activated. This results in a lower inference cost, effectively reducing the impact of trainable experts during inference.

\begin{table}[htbp]
\centering
\caption{Comparison of model complexity across different gaze estimation methods. 
"Hybrid" indicates models that combine CNN and MoE transformer and CLIP, 
while "Trans." refers to Transformer-based models. 
Parameters are reported in millions (M) and FLOPs in giga (G).}
\label{tab:complexity_compare}
\small
\setlength{\tabcolsep}{4pt}
\begin{tabular}{lccc}
\toprule
\textbf{Method} & \textbf{Params (M)} & \textbf{FLOPs (G)} & \textbf{Backbone} \\
\midrule
FullFace  \cite{zhang2017s}         & 196.60 & 2.99  & CNN \\
RT-GENE \cite{fischer2018rt}          & 82.00  & 30.81 & CNN \\
Dilated-Net \cite{chen2018appearance}      & \textbf{3.90} & 3.14  & CNN \\
Gaze360 \cite{kellnhofer2019gaze360}         & 14.60  & 12.78 & RNN \\
CA-Net \cite{cheng2020coarse}           & 34.10  & 15.60  & CNN \\
GazeTR-Pure \cite{cheng2022gaze}      & 227.3 & 58.32 & Trans. \\
Vanilla CapsNet \cite{cheng2022gaze}  & 11.80  & 2.50   & CNN \\
GazeCapsNet  \cite{muksimova2025gazecapsnet}         & 11.70  & \textbf{1.82} & CNN \\
\midrule

\textbf{Ours}             & 129.03 & 27.14 & Hybrid \\
\bottomrule
\end{tabular}
\end{table}

\mypar{Comparison of model complexity.}
We evaluate the model complexity of our GMGaze against various existing methods using different architectures. As summarized in Table~\ref{tab:complexity_compare}, while lightweight models like Dilated-Net and GazeCaps show minimal parameters and FLOPs, their performance is constrained by limited representation capacity. Conversely, transformer-based models such as GazeTR-Pure exhibit significant computational overhead (227.3M Params and 58.32 GFLOPs). Unlike these methods, by leveraging a hybrid architecture with 129.03M parameters and 27.14 GFLOPs, GMGaze achieves a reasonable trade-off. It provides a lower GE error than lightweight baselines while maintaining higher efficiency compared to transformers.

\begin{table}[ht]
\centering
\scriptsize
\setlength{\tabcolsep}{1pt}
\renewcommand{\arraystretch}{0.9}
\caption{Ablation on semantic prototype conditioning. {Frozen} keeps prototype banks fixed after CLIP-prompt initialization; {Random init} uses learnable prototypes initialized randomly.}
\label{tab:ablation_proto}
\begin{tabular}{lcccccc}
\toprule
\textbf{Setting} & \textbf{Prototypes} & \textbf{Init} & \textbf{Et} & \textbf{G} & \textbf{E} & \textbf{M} \\
\midrule
Ours & learnable & CLIP prompts & 1.44$^\circ$ & 10.16$^\circ$ & 3.22$^\circ$ & 2.49$^\circ$ \\
w/ frozen prototypes & frozen & CLIP prompts & 1.49$^\circ$ & 11.36$^\circ$ & 4.33$^\circ$ &3.41$^\circ$\\
w/ random init & learnable & random & 1.52$^\circ$ & 11.22$^\circ$ & 4.38$^\circ$&3.37$^\circ$ \\
\bottomrule
\end{tabular}
\end{table}

\mypar{Semantic prototype conditioning.}
Table~\ref{tab:ablation_proto} reports the angular error for semantic prototype conditioning. Freezing the prototypes after CLIP-prompt initialization consistently increases the error on \textbf{G/E/M} (e.g., 10.16$\to$11.36$^\circ$, 3.22$\to$4.33$^\circ$, 2.49$\to$3.41$^\circ$), indicating that prompt semantics alone are insufficient to capture dataset-specific gaze patterns and supervision signals. Moreover, making the prototypes learnable but randomly initialized further degrades performance (e.g., \textbf{E}: 3.22$\to$4.38$^\circ$), suggesting that merely introducing additional parameters does not guarantee effective modeling. Furthermore, this study does not consider techniques such as fine-tuning the entire CLIP text encoder or applying continuous prompt tuning (e.g., CoOp \cite{zhou2022learning}), since they can degrade the zero-shot ability of CLIP and introduce large GPU overhead. Thus, the proposed approach, freezing the text backbone and optimizing only the initialized continuous prototypes $\mathbf{P}$, optimally balances semantic prior exploitation with robust task-specific adaptation. Overall, learnable CLIP-initialized prototypes provide more informative contextual biases in $\boldsymbol{f}_1, \boldsymbol{f}_2$, which in turn facilitates downstream feature fusion and MoE specialization.

\mypar{Robustness under Image Degradations.}
Table~\ref{tab:robustness} reports the Mean Angular Error (MAE) of GMGaze under various common image degradations, such as Gaussian blur, low resolution, and occlusion (Cutout). Compared to the ResNet-50 baseline and GazeTR, GMGaze consistently yields the lowest MAE across all tested settings. Particularly under occlusion, GMGaze provides an MAE of 4.88$^\circ$, outperforming ResNet-50 (6.27$^\circ$) and GazeTR (5.64$^\circ$). This suggests that the robust global semantic priors provided by the pre-trained CLIP vision encoder and the adaptive expert selection mechanism enable the model to efficiently handle corrupted or incomplete input signals in real-world environments.

\begin{figure}
     \centering
\includegraphics[width=1\linewidth]{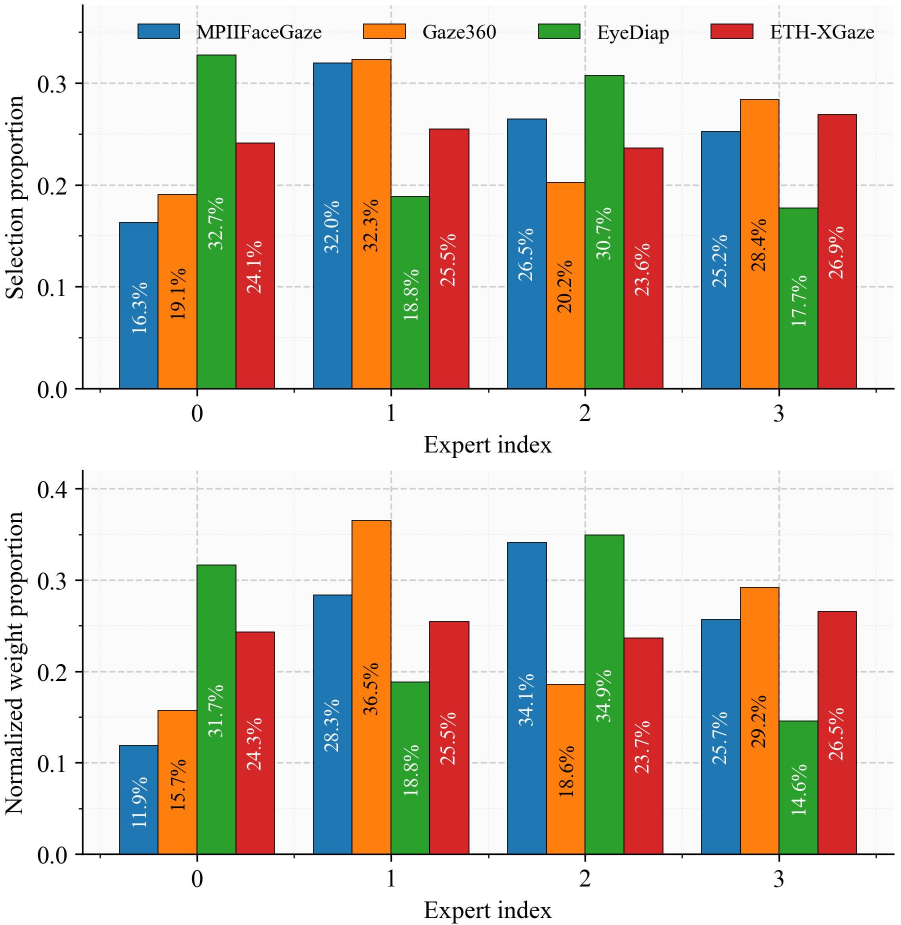}

\caption{Comparison of MoE expert load across datasets (within-domain setting). \textbf{Top}: selection proportion (percentage of token assignments per expert). \textbf{Bottom}: normalized weight proportion (sum of gating weights per expert, normalized).}
    \label{vis_load}
\end{figure}

\mypar{Module interaction} Table~\ref{tab:interaction_ablation} reports the angular error resulting from interaction configurations between SPC and MoE in GMGaze. Adding MoE alone reduces within-domain error on $\mathcal{D}_E$ (5.53$^\circ$ to 4.33$^\circ$) but decreases the performance on cross-domain ($\mathcal{D}_{Et} \to \mathcal{D}_E$, error 9.96$^\circ$), indicating MoE overfits to domain-specific features without guidance. Furthermore, the use of SPC improves cross-domain generalization (9.07$^\circ$ to 8.61$^\circ$) but slightly degrades within-domain performance. Finally, combing SPC with MoE achieves the best performance: 3.22$^\circ$ (within-domain) and 7.99$^\circ$ (cross-domain), confirming SPC acts as a semantic anchor to prevent MoE overfitting and enable generalized gaze representations.

\mypar{MoE module.} Table~\ref{tab:ablation2} reports the angular error for with and without MoE module. The use of MoE leads to a lower angular error compared to without MoE (e.g., 2.49$^\circ$ vs. 4.20$^\circ$). These results highlight the potential of using MoE in transformer model. Figure \ref{vis_load} illustrates the expert utilization across the four datasets.

\begin{figure}
     \centering
\includegraphics[width=1\linewidth]{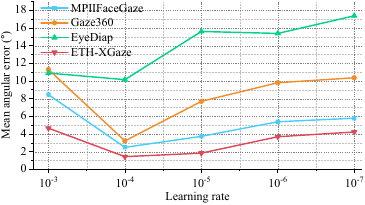}

\caption{Ablation study of learning rate's effect on mean angular error across datasets M, E, G, and Et.}
    \label{LRCURVE}
\end{figure}

\mypar{Learning rates.} Figure \ref{LRCURVE} shows the error rates with various learning rates. Setting the learning rate to $10^{-4}$ leads to the best performance. Both larger and smaller learning rates reduce the performance.

\mypar{Feature separation loss.} As reported in Table \ref{tab:sep}, the feature separation loss reduces the average angular errors of the four cross-domain transfer routes. For example, the angular error on $\mathcal{D}_{Et} \to \mathcal{D}_M$ decreases from 8.42$^\circ$ to 4.81$^\circ$, and on $\mathcal{D}_{G} \to \mathcal{D}_E$, the angular error decreases from 9.57$^\circ$ to 6.90$^\circ$. These results suggest that the feature separation loss effectively reduces the correlation between $\boldsymbol{f}_1$ and $\boldsymbol{f}_2$, enabling prototype conditioning to provide more complementary information flow. Consequently, it assigns clearer signals to MoE experts during the domain alignment and routing stages, thereby enhancing cross-domain generalization ability.

\begin{table}[htbp]
\centering
\caption{Robustness evaluation under various synthetic image degradations on the MPIIFaceGaze dataset. Results are reported in Mean Angular Error (MAE) ($^\circ$).}
\label{tab:robustness}
\resizebox{\columnwidth}{!}{
\begin{tabular}{lcccc}
\toprule
\textbf{Model} & \textbf{Clean Image} & \textbf{Gaussian Blur} & \textbf{Low Resolution} & \textbf{Occlusion (Cutout)} \\
\midrule
ResNet-50 \citep{zhang2017s} & 4.12$^\circ$ & 4.87$^\circ$ & 5.62$^\circ$ & 6.27$^\circ$ \\
GazeTR$^\circ$ \citep{cheng2022gaze}       & 4.00$^\circ$ & 4.60$^\circ$ & 5.08$^\circ$ & 5.64$^\circ$ \\
\rowcolor{gray!15} 
\textbf{GMGaze (Ours)}       & \textbf{2.49$^\circ$} & 3.19$^\circ$ & 3.75$^\circ$ & 4.88$^\circ$ \\
\bottomrule
\end{tabular}
}
\end{table}

\begin{table}[ht]

    \centering\scriptsize

    \setlength{\tabcolsep}{5pt}

    \renewcommand{\arraystretch}{1.1}

    \caption{Analysis of module interactions between Semantic Prototype Conditioning (SPC) and Mixture of Experts (MoE) on both within-domain ($\mathcal{D}_E$) and cross-domain ($\mathcal{D}_{Et} \to \mathcal{D}_E$) setups.}

    \label{tab:interaction_ablation}

    \begin{tabular}{ccccc}

        \toprule

        \multicolumn{2}{c}{\textbf{Modules}} & & \multicolumn{2}{c}{\textbf{Angular Error ($^\circ$)}} \\

        \cmidrule(lr){1-2} \cmidrule(lr){4-5}

        \textbf{SPC} & \textbf{MoE} & & \textbf{Within ($\mathcal{D}_E$)} & \textbf{Cross ($\mathcal{D}_{Et} \to \mathcal{D}_E$)} \\

        \midrule

        $\times$ & $\times$ & Baseline & 5.53$^\circ$ & 9.07$^\circ$ \\
        $\checkmark$ & $\times$ & +SPC only & 5.78$^\circ$ & 8.61$^\circ$ \\ 
        $\times$ & $\checkmark$ & +MoE only & 4.33$^\circ$ & 9.96$^\circ$ \\
        $\checkmark$ & $\checkmark$ & \textbf{GMGaze} & \textbf{3.22$^\circ$} & \textbf{7.99$^\circ$} \\

        \bottomrule

    \end{tabular}

\end{table}

\begin{table}[ht]
    \centering\scriptsize
    \setlength{\tabcolsep}{6pt}
    \renewcommand{\arraystretch}{0.9}
    \caption{Angular errors of with MoE and without MoE.}
    \label{tab:ablation2}
    \begin{tabular}{lcccc}
        \toprule
        \multirow{2}{*}{\textbf{MoE}} & \multicolumn{4}{c}{\textbf{Datasets}} \\
        \cmidrule(lr){2-5}
        & \textbf{Et} & \textbf{G} & \textbf{E} & \textbf{M} \\
        \midrule
        $+$ & 1.44$^\circ$ & 10.16$^\circ$ & 3.22$^\circ$ & 2.49$^\circ$ \\
        w/o & 4.38$^\circ$ & 10.72$^\circ$ & 5.78$^\circ$ & 4.20$^\circ$ \\
        \bottomrule
    \end{tabular}
\end{table}

\begin{table}[ht]
    \centering\scriptsize
    \setlength{\tabcolsep}{6pt}
    \renewcommand{\arraystretch}{1}
    \caption{Test angular error under cross-domain setting for with and without feature seperation loss.}
    \label{tab:sep}
    \begin{tabular}{lcccc}
        \toprule
        \multirow{2}{*}{ \textbf{}}&\multicolumn{4}{c}{\textbf{Datasets}} \\
        \cmidrule(lr){2-5}
        & $\mathcal{D}_{Et} \to \mathcal{D}_M$ & $\mathcal{D}_{Et} \to \mathcal{D}_E$ & $\mathcal{D}_{G} \to \mathcal{D}_M$ & $\mathcal{D}_{G} \to \mathcal{D}_E$ \\
        \midrule
        $+$ & 4.81$^\circ$ & 7.99$^\circ$ & 5.53$^\circ$ & 6.90$^\circ$ \\
        w/o &8.42$^\circ$ &9.96$^\circ$&7.03$^\circ$ &9.57$^\circ$ \\
        \bottomrule
    \end{tabular}
\end{table}

\section{Discussion}
\label{discussion}

\mypar{Comparing with SOTA.} Based on the results as summarized in Table \ref{tab:gaze_comparison} and Table \ref{tab:gaze_crossdomain}, we observe that coupling CLIP-aligned prototype conditioning with unified multi-scale token fusion and a MoE module learns more generalizable appearance-based gaze estimation under illumination, pose, and background shift. Notably, it achieves 2.49$^\circ$, 3.22$^\circ$, 10.16$^\circ$, and 1.44$^\circ$ on datasets \textbf{M}, \textbf{E}, \textbf{G}, and \textbf{Et} respectively. Furthermore, ablation studies confirm that the performance gains are not simply relying on parameter scaling, as prototype-conditioned global tokens alone remain weakly constrained (errors above 7$^\circ$, 10$^\circ$, 28$^\circ$, and 10$^\circ$ on the four datasets), indicating that coarse semantic priors without fine spatial structure fail to capture critical periocular micro-texture and shading cues. Introducing high-resolution CNN tokens reduces this gap (e.g., \textbf{M}, from 7.66$^\circ$ to 3.20$^\circ$), and the use of CLIP patch tokens delivers a further refinement (\textbf{M}, from 3.20$^\circ$ to 2.49$^\circ$; \textbf{E}, from 4.39$^\circ$ to 3.22$^\circ$), suggesting that low-level texture, mid-level semantic structure, and prototype-guided context are complementary when co-attended within a single sequence space rather than fused late.

\mypar{Fusion and prototype ablations.}
The results as reported in Table~\ref{tab:ablation_fusion} and Table~\ref{tab:ablation_proto} further support the proposed designs. For example, early fusion (Table~\ref{tab:ablation_fusion}) provides the best performance across all four datasets, indicating that injecting the two context-biased tokens early is more effective than late fusion via cross-attention or merging only at the prediction stage.This is consistent with the goal of using the global context to guide the generation of local tokens within the Transformer. Furthermore, prototype conditioning benefits from both {CLIP-aligned initialization} and {end-to-end adaptation} (Table~\ref{tab:ablation_proto}): freezing prototypes or using random initialization degrades performance, especially on diverse datasets (G/E/M), suggesting that prompt semantics provide a useful prior but require to be tuned to dataset-specific patterns under gaze supervision.

\mypar{Simple FFN vs. MoE.} Removing MoE produces uniform performance degradation (e.g., \textbf{Et}, from 1.44$^\circ$ to 4.38$^\circ$), suggesting that a single dense FFN underfits rare, illumination-perturbed or high-yaw sub-distributions and overfits dominant frontal well-lit modes. Despite the remarkable performance of the MoE, it introduces practical trade-offs: routing variance and potential tail latency under unbalanced expert loads, and sensitivity to the load-balancing coefficient (if mis-tuned, reduces capacity where extreme yaw or deep shadow samples would benefit). Furthermore, the discrete prototype selection used in our study (argmax with learnable temperature) helps sharpen semantic routing and avoid noisy soft mixtures. However, it imposes a finite granularity that may under-represent continuous illumination gradients or composite mixed‑spectrum indoor–outdoor transitions.

\mypar{Feature separation loss.} Introducing the feature separation loss (Eq.~\ref{eq:sep_loss}) consistently reduces the correlation between the two global vectors and improves cross‑domain routing clarity (Table~\ref{tab:sep}), suggesting the loss helps prototypes provide more complementary signals to MoE experts; nevertheless, its hyperparameter sensitivity and the extent to which it mitigates source bias require further sensitivity analysis.



\begin{figure}
     \centering
\includegraphics[width=1\linewidth]{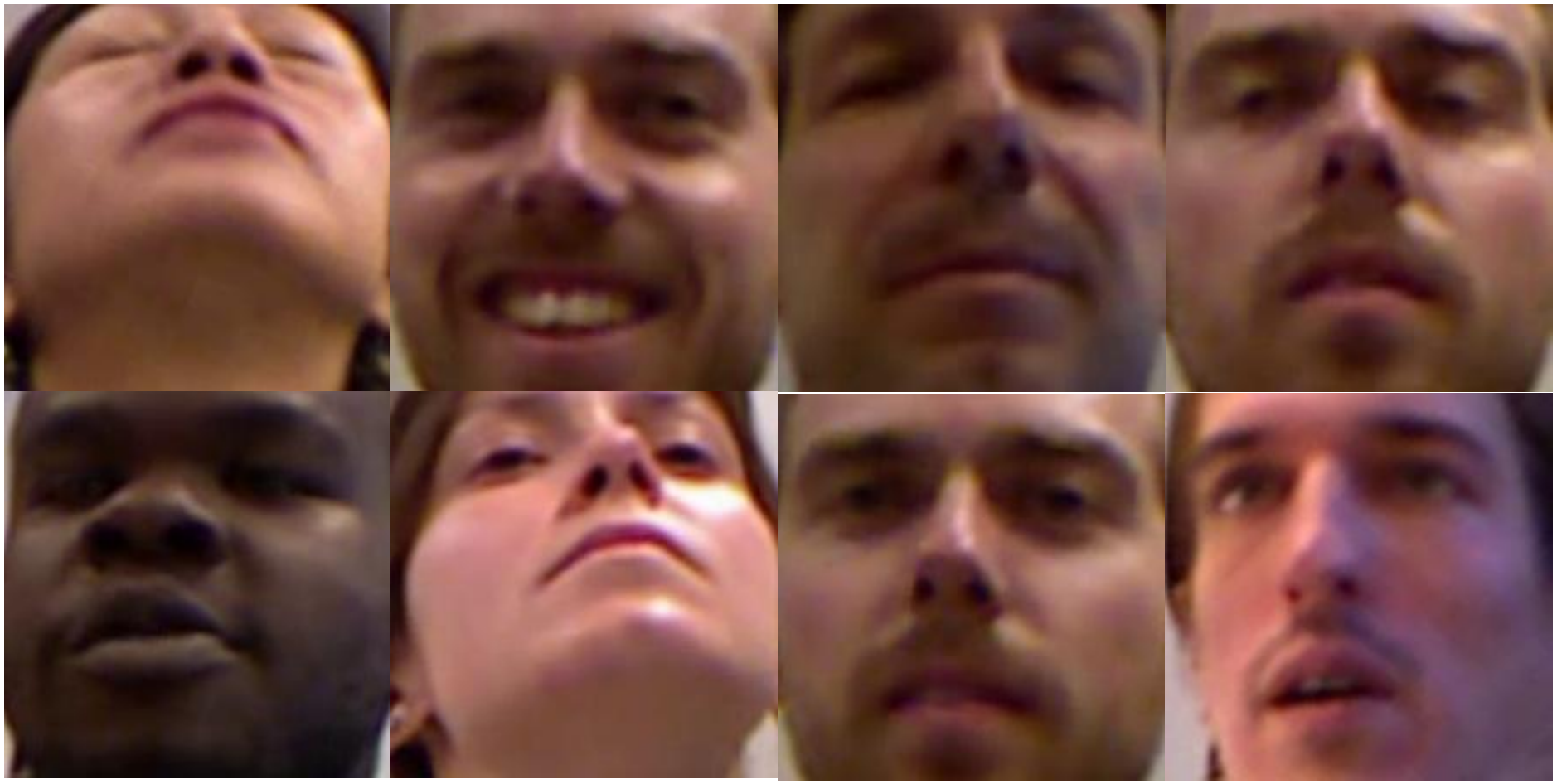}\\
\caption{Example of failure cases from cross-domain setting using the EYEDIAP as target domain.}
    \label{faliure_cas}
\end{figure}
\mypar{Cross-domain.} Table~\ref{tab:gaze_crossdomain} shows that improvements are selective: the proposed model achieves lowest errors on $\mathcal{D}_{Et}\!\to\!\mathcal{D}_M$ (4.81$^\circ$) and $\mathcal{D}_{G}\!\to\!\mathcal{D}_M$ (5.53$^\circ$), while it yields comparable or worse metrics on $\mathcal{D}_{G}\!\to\!\mathcal{D}_E$ (6.90$^\circ$ vs.\ 6.49$^\circ$ with CRGA) and $\mathcal{D}_{Et}\!\to\!\mathcal{D}_E$ (7.99$^\circ$ vs.\ 5.10$^\circ$ with RUDA). These mixed results indicate that the proposed model are robust when adapting into the more heterogeneous target $\mathcal{D}_M$, yet adapts less effectively when the target is the more controlled $\mathcal{D}_E$, where a narrower appearance distribution appears to expose over‑specialization of prototypes learned on $\mathcal{D}_{Et}$ or $\mathcal{D}_{G}$. The larger gap on $\mathcal{D}_{Et}\!\to\!\mathcal{D}_E$ suggests that argmax prototype selection with adversarial adaptation is insufficient to narrow the distribution shifts between a wide‑pose, illumination‑rich source and a low‑diversity, lab‑style target. 

\mypar{Analysis of failure cases.} 
As illustrated in Figure~\ref{faliure_cas}, representative failure cases under cross-domain setting (target domain: $\mathcal{D}_E$) reveal the boundary conditions of our semantic prototype conditioning. Qualitatively, prediction errors are predominantly concentrated in samples characterized by extreme upward pitch angles or insufficient facial illumination where ocular details are heavily obscured. In cases involving extreme pitch, the periocular morphology exhibits geometric distortion that occasionally exceeds the range of representation of the proposed pre-defined pose prototypes. 
As shown in Figure \ref{data_distribution}, while benchmarks like ETH-XGaze and Gaze360 cover diverse continuous pose spaces (e.g., yaw angles spanning up to $\pm 80^\circ$), the $\mathcal{D}_E$ distribution is densely clustered within a narrow, rectangular manifold with limited yaw and predominantly positive pitch. Thus, models trained on the highly generalized manifolds of $\mathcal{D}_{Et}$ or $\mathcal{D}_G$ struggle to adapt their predictions for the intensely concentrated, micro-variational data manifold of $\mathcal{D}_E$. This suggests that incorporating adaptive soft-assignment or continuous prompt-tuning could further improve the flexibility of the model when handling transitions into such constrained and degraded domains.

\section{Conclusion}
\label{conclusion}
We proposed GMGaze, which leverages two complementary context-biased tokens, early fusion of semantic, CLIP patch and CNN tokens, and MoE layers. Experiments on four benchmarks and ablations show that this design is particularly strong at handling challenging cases, and that the gains come from how capacity is allocated across tokens rather than from a simple increase in dense parameters. At the same time, the method has clear weaknesses. The prototype banks are initialized from hand-crafted CLIP prompts and the routing remains discrete, which restricts the granularity of context modeling and may miss subtle or mixed conditions. While the prompt vocabulary is fixed, the prototype vectors are learnable and are updated during training; however, hard selection updates only the chosen prototypes per iteration and may under-represent continuous context variations (e.g., smooth illumination gradients). In addition, the current model is single‑frame and uses a relatively simple adversarial alignment strategy, which helps on some cross‑domain routes but does not outperform specialized methods.

Despite these limitations, the framework introduced here can be reused by others in the field. The way we turn text prompts into learnable prototype banks, build a unified token sequence fusing semantic, CLIP and CNN tokens, and insert MoE layers can be directly transferred to related problems such as head pose estimation, driver monitoring or attention analysis without redesigning visual backbones. The feature separation loss and adversarial training setup are also generic and can serve as simple regularizers in other domain adaptation frameworks.

Future work will focus on three directions: (i) learning data-driven, dynamic prototypes that automatically adapt to new domains and devices; (ii) extending GMGaze to video with temporal attention and sequence-level routing to exploit motion and temporal consistency; and (iii) compressing the multi-expert backbone via expert distillation and pruning to meet the latency and memory constraints of VR/AR headsets and mobile platforms.





\section*{Acknowledgments}
This research was funded by the Guangxi Science and Technology Base and Talent Project (2022AC18004, 2022AC21040), and the National Natural Science Foundation of China grant number 82260360. 

{
\bibliographystyle{unsrt}
\bibliography{cas-refs}
}

\end{document}